\documentclass{article}

\usepackage{graphicx}
\usepackage{amsmath}
\usepackage{amssymb}
\usepackage{booktabs}
\usepackage{adjustbox}
\usepackage{capt-of}
\usepackage{placeins}
\usepackage[table]{xcolor}
\usepackage{titlesec}
\usepackage[final]{corl_2026} 
\usepackage{sharpa_branding}

\SharpaBrandingOn

\titlespacing*{\section}{0pt}{0.15ex plus 0.03ex minus 0.03ex}{0.06ex plus 0.02ex}
\titlespacing*{\subsection}{0pt}{0.12ex plus 0.03ex minus 0.03ex}{0.04ex plus 0.01ex}
\titlespacing*{\subsubsection}{0pt}{0.08ex plus 0.02ex minus 0.02ex}{0.03ex plus 0.01ex}
\newsavebox{\zaxistablebox}
\newsavebox{\yaxistablebox}
\newsavebox{\xaxistablebox}
\newsavebox{\unseentablebox}
\newsavebox{\wsmcontroltablebox}

\definecolor{wmcraftcolor}{RGB}{0,100,150}
\definecolor{deeppurple}{RGB}{90, 0, 120}
\definecolor{darkgreen}{RGB}{0, 100, 60}
\definecolor{tealpro}{RGB}{0, 120, 130}
\definecolor{wmcraftcolorv2}{RGB}{0,70,140}
\definecolor{royalblue}{RGB}{0, 90, 180}
\definecolor{navyblue}{RGB}{0, 0, 128}
\definecolor{cyanblue}{RGB}{0, 150, 180}
\definecolor{colorFst}{RGB}{214, 238, 205}
\definecolor{colorSnd}{RGB}{235, 246, 218}
\definecolor{colorTrd}{RGB}{255, 248, 205}

\newcommand{\fs}{\cellcolor{colorFst}}   
\newcommand{\nd}{\cellcolor{colorSnd}}      
\newcommand{\rd}{\cellcolor{colorTrd}}      

\title{\textcolor{cyanblue}{WM-Craftnet}: World Synesthesia Model for Generalizable and Robust Dexterous In-Hand Manipulation}

\author{
Jie Yin$^{\dagger}$, Zeyuan Zhao$^{\dagger,\ddagger}$, Xiaojing Tan, Yang Liu$^{\ddagger}$, Chiyu Wang, Xinyang Gu$^{*}$\\
Sharpa Robotics\\
\href{https://wmcraftnet.github.io}{\textcolor{cyanblue}{\textbf{{https://wmcraftnet.github.io}}}}
}
\hypersetup{
  pdftitle={WM-Craftnet: World Synesthesia Model for Generalizable and Robust Dexterous In-Hand Manipulation},
  pdfauthor={Jie Yin, Zeyuan Zhao, Xiaojing Tan, Yang Liu, Chiyu Wang, Xinyang Gu}
}

\begin{document}
\maketitle
\makeatletter
\begingroup
\renewcommand{\thefootnote}{}
\long\def\@makefntext#1{\noindent#1}
\footnotetext{\(^{\dagger}\) are co-first authors.\quad
\(^{\ddagger}\) contributed while interning at Sharpa Robotics.\\
\(^{*}\) Corresponding author: Xinyang Gu (\href{mailto:zlw21gxy@gmail.com}{\texttt{zlw21gxy@gmail.com}}).}
\endgroup
\makeatother
\setcounter{footnote}{0}
\vspace{-10mm}

\begin{abstract}
Generalizable and robust dexterous in-hand manipulation requires a policy to infer object pose, geometry, contact, and potential slip from partial and noisy observations. Although recent tactile and visuotactile RL methods achieve strong in-hand rotation in controlled settings, their robustness often degrades under pose shifts, force disturbances, and object variation. We propose WM-Craftnet, a world-model-conditioned framework that learns compact action-conditioned latent dynamics from proprioception, depth, tactile sensing, and actions, supervised by multimodal reconstruction and reward prediction. Rather than using the world model for latent imagination or policy optimization, WM-Craftnet uses the learned World Synesthesia Model (WSM) as recurrent task context for an asymmetric actor--critic policy. Importantly, WSM is trained to reconstruct clean depth targets from noisy depth inputs, providing a denoised geometric state for real-robot deployment. Ablations over recurrent baselines, auxiliary heads, tactile masking, and WSM modality heads show that predictive world modeling, clean-depth supervision, and tactile contact cues all shape the learned state. A WSM pretrained on nine \(z\)-axis objects serves as a reusable prior for \(49\)-object downstream policy learning. This context improves multi-object rotation, with quantitative and qualitative evidence for unseen-object, perturbation-recovery, and sim-to-real transfer.
\end{abstract}


\section{Introduction}
\label{sec:intro}

Dexterous in-hand manipulation is essential for contact-rich skills such as object reorientation, tool use, and fine-grained adjustment. Unlike arm-gripper manipulation with relatively stable contacts, in-hand manipulation must control a partially observed object through intermittent fingertip interactions while adapting to changes in pose, geometry, and external force.

Learning-based methods have made substantial progress with proprioception, touch, and vision \citep{yin2023touchdexterity,qi2023rotateit,handa2023dextreme,qi2023hand,yang2024anyrotate}. However, policies trained on narrow object sets or fixed initial poses can overfit to object-specific finger gaits, making them fragile to pose offsets, force disturbances, and object drift. Some methods add object identities or shape embeddings, but this ties deployment to known annotations and limits general-purpose manipulation.

Pose-robust adjustment requires inferring object geometry, pose, contact state, and slip risk under partial observability. Depth and touch provide complementary evidence, but simply concatenating sensor streams leaves the policy to learn temporal contact dynamics from noisy fused observations, especially when real wrist-depth inputs are corrupted or shifted from clean simulation.

Our goal is to study how an action-conditioned latent state can improve task-relevant physical inference for contact-rich manipulation. Inspired by human synesthetic physical reasoning, we combine global geometry from vision, local contact evidence from touch, and action history that reveals how contacts evolve under control. Prior visuotactile work fuses these cues for in-hand control \citep{yuan2024robot}, and world-model representations can provide deployable predictive features \citep{lai2024wmp}. WM-Craftnet uses this idea to anticipate contact transitions, object drift, and slip, enabling adjustment under perturbations rather than repetition of a fixed gait.

WM-Craftnet learns a World Synesthesia Model with a Dreamer-style recurrent state from proprioception, noisy depth, tactile observations, and recent actions, with reward used as a prediction target during training. The policy is conditioned on this recurrent feature, which captures object state, geometry, and contact evolution. During WSM training, noisy augmented depth is used as input while clean wrist-depth targets supervise reconstruction, encouraging denoised geometry for downstream control. This representation improves robust multi-object rotation, supports unseen-object evaluation, and enables WSM pretraining as a reusable prior for downstream \(49\)-object control. We summarize our contributions as follows:

\begin{itemize}
    \item We propose WM-Craftnet, a World Synesthesia Model-based framework that learns an action-conditioned recurrent representation from proprioception, noisy depth, tactile sensing, actions, and multimodal prediction targets for contact-rich dexterous manipulation.
    \item We analyze WSM with recurrent-history baselines, auxiliary-head and modality-head ablations, latent-state diagnostics, clean-depth reconstruction, and pretraining, showing that the predictive representation can serve as a reusable prior for \(49\)-object downstream control.
    \item We conduct quantitative simulation and real-world evaluations, supported by qualitative videos, showing that WSM improves single-policy manipulation across object variations, unseen objects, perturbations, and enables sim-to-real transfer on a human-sized \(22\)-DoF five-finger dexterous hand.
\end{itemize}
\section{Related Work}
\label{sec:related}

\paragraph{Learning-based in-hand manipulation.}
Learning-based systems have achieved sim-to-real in-hand control through several complementary strategies. Rapid Motor Adaptation infers object properties online from proprioceptive history for \(z\)-axis rotation \citep{qi2023hand}, whereas DeXtreme couples a domain-randomized reorientation policy with a vision-based pose estimator \citep{handa2023dextreme}. Touch Dexterity demonstrates vision-free rotation with distributed binary contact sensing \citep{yin2023touchdexterity}, and AnyRotate uses dense tactile feedback for gravity-invariant multi-axis rotation \citep{yang2024anyrotate}. More complex reorientation can compose pretrained axis-wise skills \citep{qi2025simple} or route among shape-specialized experts \citep{wan2025dexremoe}. Other approaches adapt with real pen-spinning trajectories \citep{wang2024lessons}, learn joint-wise actuator corrections from hardware data \citep{liu2025dexndm}, or randomize 3D Gaussians for monocular pose estimation \citep{bhardwaj2026viserdex}. In adjacent dexterous grasping, Goal-Auxiliary Actor-Critic combines point-cloud control with an auxiliary goal objective \citep{wang2022goalaux}, while DextrAH-G couples pixel policies with geometric-fabric control priors \citep{lum2024dextrah}. These methods address grasp acquisition; WM-Craftnet instead learns recurrent interaction state for sustained in-hand control.

\paragraph{Visuotactile perception for dexterity.}
RotateIt distills privileged object shape and physical properties into a temporal transformer driven by vision, touch, proprioception, and action history \citep{qi2023rotateit}. Robot Synesthesia projects camera, hand, and active tactile points into a common 3D representation before teacher--student policy distillation \citep{yuan2024robot}. NeuralFeels fuses vision and touch in an online neural field for object-state estimation \citep{suresh2024neuralfeels}, while tactile skins with normal and shear sensing enable feedback-driven in-hand translation \citep{yin2025learning}. ViTacFormer learns cross-modal features for visuotactile dexterous manipulation \citep{heng2025vitacformer}; image-based hand-object reconstruction also shows how contact-aware geometric objectives can structure a latent representation \citep{hasson2019joint}. WM-Craftnet differs by learning action-conditioned recurrent dynamics with multimodal prediction objectives and exposing the deterministic state as detached PPO context, rather than regressing a fixed privileged encoding or separating perception from control.

\paragraph{World models for robot control.}
World Models introduced compact learned dynamics for control \citep{ha2018worldmodels}, and Dreamer optimizes behavior through trajectories imagined in a recurrent state-space model \citep{hafner2019dreamer,hafner2023dreamerv3}. DayDreamer applies this paradigm directly to physical robots \citep{wu2022daydreamer}, while MoDem-V2 uses visual world models for real-world contact-rich manipulation \citep{lancaster2023modemv2}. Robotic World Model instead treats learned dynamics as a neural simulator for policy optimization \citep{li2025roboticworldmodel}. In the representation-conditioning branch, World Model-based Perception supplies a stopped-gradient recurrent state to a locomotion policy without imagined PPO rollouts \citep{lai2024wmp}, and DreamMimic uses world-model predictive features for visuomotor whole-body policy distillation \citep{yin2026dreammimic}. WM-Craftnet extends this branch to visuotactile in-hand manipulation, adding noisy-to-clean depth reconstruction and reuse of the predictive model across object distributions.


\section{WM-Craftnet Framework}
\label{sec:method}
\begin{figure}[t]
  \centering
  \begin{adjustbox}{width=0.95\linewidth,center}
    \includegraphics{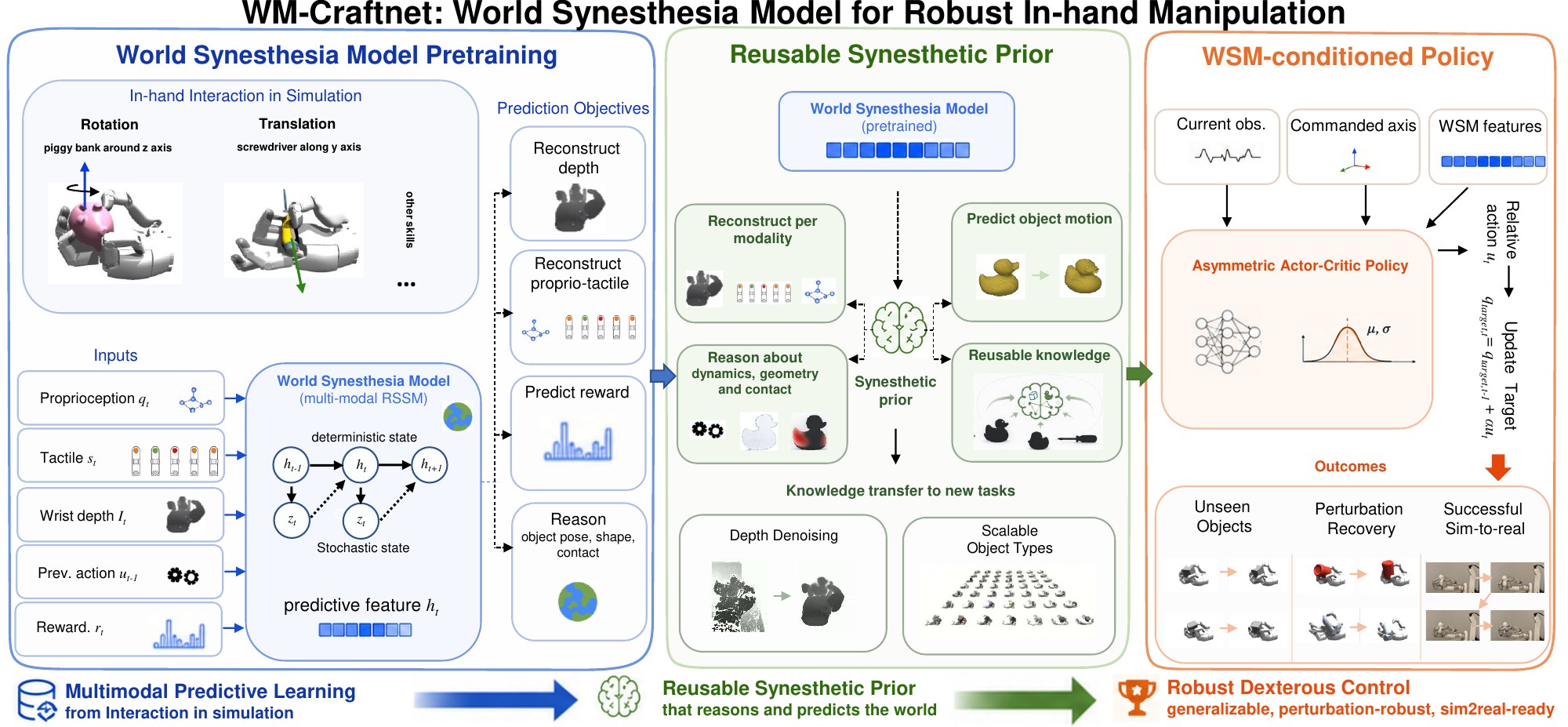}
  \end{adjustbox}
  \caption{Overview of WM-Craftnet. A Dreamer-style recurrent state-space model (RSSM) learns a denoised recurrent state from multimodal observations and provides it as predictive context for the manipulation policy.}
  \label{fig:wmcraft-pipeline}
\end{figure}

WM-Craftnet targets partially observed in-hand manipulation on a human-sized \(22\)-DoF five-finger hand. We define the \emph{World Synesthesia Model} as a Dreamer-style recurrent latent dynamics model trained with multimodal reconstruction and reward prediction, whose detached deterministic state serves as deployable policy context. The name reflects its cross-modal predictive role: visual geometry, tactile contact, proprioception, and action history jointly constrain the latent interaction state. This predictive state supports object-ID-free adaptation from deployable visuotactile observations and serves as a recurrent state estimate for the actor, as illustrated in Figure~\ref{fig:wmcraft-pipeline}.

\vspace{0mm}
\subsection{Problem Formulation and Observations}
We consider continuous in-hand rotation of a grasped object around a commanded axis \(\mathbf{a}\in\{\pm x,\pm y,\pm z\}\). At time \(t\), we denote the deployable policy observation by
\begin{equation}
\label{eq:policy-obs}
\mathbf{o}_t=\big(\mathbf{p}^{\text{stack}}_t,\mathbf{c}^{\text{stack}}_t,\mathbf{D}_t,\mathbf{u}_{t-1}\big),
\end{equation}
where \(\mathbf{p}^{\text{stack}}_t\) and \(\mathbf{c}^{\text{stack}}_t\) are short stacks of normalized hand proprioception and binary tactile/contact measurements, \(\mathbf{D}_t\) is the current noisy wrist-depth stream, and \(\mathbf{u}_{t-1}\) is the previous joint-target command. Importantly, \(\mathbf{o}_t\) contains only on-robot sensor streams and control history; it does not include clean simulator depth or any depth reconstructed by the WSM decoder. The actor conditions on \(\mathbf{o}_t\), the commanded axis \(\mathbf{a}\), and the projected WSM feature \(\psi(\mathbf{w}_t)\). The policy outputs a relative \(22\)-DoF joint target command \(\mathbf{u}_t\), which is smoothed and executed by the low-level hand controller.

The objective is signed object rotation around the target axis while maintaining a stable grasp. Let \(\mathbf{R}^{o}_t\in SO(3)\) denote object orientation. One-step progress is
\begin{equation}
\Delta \theta_t
=
\mathrm{proj}_{\mathbf{a}}\!\left(
\log\!\left(\mathbf{R}^{o}_{t+1}(\mathbf{R}^{o}_{t})^{-1}\right)
\right),
\end{equation}
where \(\log(\cdot)\) maps rotation to axis-angle form and \(\mathrm{proj}_{\mathbf{a}}\) projects onto the commanded axis. During simulation training, object pose, velocity, and contact-force fields support reward computation, critic learning, and diagnostics. Table~\ref{tab:obs-action-interface} lists the complete observation/action interface.

\subsection{Multimodal Predictive State Encoder}
The encoder maps sensory streams to a latent state for prediction and control. The world-model update receives a single-step proprioceptive vector \(\mathbf{p}_t\), tactile/contact vector \(\mathbf{c}_t\), noisy normalized depth image \(\mathbf{I}_t\), previous action \(\mathbf{u}_{t-1}\), and episode-start flag \(m_t\). The rollout record additionally contains clean depth and reward as training targets; neither is required as a deployment-time input. The actor observes stacked proprioceptive/tactile state, whereas the world model uses unstacked low-dimensional vectors to keep dynamics compact.

Given rollout data
\begin{equation}
\mathcal{D}
=
\{(\mathbf{p}_t,\mathbf{c}_t,\mathbf{I}_t,\mathbf{I}^{\mathrm{clean}}_t,\mathbf{u}_{t-1},r_t,m_t)\}_{t=1}^{T},
\end{equation}
the proprioceptive and tactile/contact streams are encoded with MLP branches and the depth stream is encoded with a CNN branch. These features are fused by a Dreamer-style multimodal encoder:
\begin{equation}
\mathbf{e}_t=f_{\text{enc}}(\mathbf{p}_t,\mathbf{c}_t,\mathbf{I}_t).
\end{equation}
Depth is clipped and cropped around the hand-object workspace. WSM input depth is augmented with temporally correlated dropout, Gaussian noise, and small random rotations, while the reconstruction target remains the clean crop-only ground-truth depth. During deployment, the actor uses the noisy sensor stream together with the recurrent WSM feature. WSM is trained to reconstruct clean hand-object geometry from corrupted observations, while its recurrent dynamics provide temporal memory from a compact per-step encoder embedding.

\subsection{World Model for In-Hand Manipulation}
WM-Craftnet uses a Dreamer-style recurrent state-space model (RSSM) as its world model. World-model optimization is decoupled from the PPO gradient path, and the actor receives a detached deterministic latent feature. Let \(\mathbf{z}_t\) be the stochastic latent state and \(\mathbf{h}_t\) be the deterministic recurrent state. Given the previous latent state, previous action, and current encoded observation, the posterior update is
\begin{equation}
\mathbf{z}_t,\mathbf{h}_t
=
f_{\text{wm}}\!\left(\mathbf{z}_{t-1},\mathbf{h}_{t-1},\mathbf{u}_{t-1},\mathbf{e}_t,m_t\right).
\end{equation}
The actor-side feature is the deterministic component
\begin{equation}
\label{eq:wm-feature}
\mathbf{w}_t=\mathbf{h}_t,
\end{equation}
which is detached before entering the policy network. In the current implementation, \(512\)-dim \(\mathbf{h}_t\) is projected by a small MLP to a \(16\)-dim feature before being concatenated with the policy observation.

The world model is trained from a replay buffer populated during PPO rollouts. Each row stores \((\mathbf{p}_t,\mathbf{c}_t,\mathbf{I}_t,\mathbf{I}^{\mathrm{clean}}_t,\mathbf{u}_{t-1},r_t,m_t)\), where \(\mathbf{I}_t\) is the noisy/augmented depth input and \(\mathbf{I}^{\mathrm{clean}}_t\) is the clean depth target used by the decoder. Training samples contiguous chunks and optimizes the Dreamer-style model objective:
\begin{equation}
\mathcal{L}_{\text{wm}}
=
\mathcal{L}_{\text{img}}
+ \mathcal{L}_{\text{prop}}
+ \lambda_c \mathcal{L}_{\text{tac}}
+ \lambda_r \mathcal{L}_{\text{reward}}
+ \lambda_{\text{dyn}}\mathcal{L}_{\text{dyn-KL}}
+ \lambda_{\text{rep}}\mathcal{L}_{\text{rep-KL}}
+ \mathcal{L}_{\text{aux}}.
\end{equation}
Here \(\mathcal{L}_{\text{img}}\) reconstructs \(\mathbf{I}^{\mathrm{clean}}_t\) from the latent state inferred with noisy depth input, \(\mathcal{L}_{\text{prop}}\) reconstructs proprioception, \(\mathcal{L}_{\text{reward}}\) predicts rollout reward, and the KL terms regularize posterior/prior dynamics. Reward supervises the predictive model during training. Together, these targets encourage a geometry- and contact-aware state for deployment with noisy wrist-depth observations.
Inspired by recent rotation and grasping systems \citep{suresh2024neuralfeels,hasson2019joint,wang2022goalaux,heng2025vitacformer,lum2024dextrah}, the full WSM uses depth, proprioception, tactile contact, reward, object pose, value, and object shape auxiliary heads to learn a task-relevant predictive state. During simulation training, the pose and shape heads use simulator labels, with shape represented by a basis-point-set-to-mesh displacement vector. Appendix~\ref{app:wsm-mechanisms} analyzes the contribution of these heads.

\subsection{Pretrained Synesthetic Prior for Multi-Object Control}
Scaling one dexterous policy from a small object set to many new objects requires inferring how geometry, contact, slip, and action-conditioned motion interact. WM-Craftnet addresses this by making WSM reusable across object distributions. After pretraining on contact-rich data from the nine-object \(z\)-axis setting, the predictive model initializes the \(49\)-object experiment and continues adapting to new rollouts, while a task-specific controller is learned with reinforcement learning.

The resulting world model acts as a reusable physical prior. It captures how depth, proprioception, action, reward, object motion, and contact evidence co-vary, giving downstream PPO a structured recurrent representation of contact transitions, object motion, and slip.
\subsection{World Model-Conditioned Manipulation Policy}

The actor receives the current policy observation, commanded axis, and projected World Synesthesia Model feature:
\begin{equation}
\mathbf{u}_t
=
\pi_{\theta}\!\left(
\mathbf{o}_t,\mathbf{a},\psi(\mathbf{w}_t)
\right),
\end{equation}
where \(\mathbf{o}_t\) is the deployable observation in Eq.~\eqref{eq:policy-obs} and \(\psi(\mathbf{w}_t)\) is the projected deterministic WSM feature derived from Eq.~\eqref{eq:wm-feature}.
Actions are relative joint target updates,
\begin{equation}
\mathbf{q}^{\text{target}}_t
=
\mathbf{q}^{\text{target}}_{t-1}
+ \alpha \mathbf{u}_t,
\end{equation}
followed by signal-level smoothing and hardware-limit clamping.

The policy is trained with PPO in simulation using the same reward for all baselines. The reward encourages stable, human-like, and goal-consistent object rotation while suppressing off-axis motion, object drift, excessive effort, large actions, and unstable contacts; Appendix~\ref{app:policy-training} gives the full reward expression, reset rules, and weights.

The actor input therefore comprises deployable observations \(\mathbf{o}_t\) and the WSM feature \(\psi(\mathbf{w}_t)\), while the critic additionally receives simulator state during training. After each PPO epoch, replay chunks update the RSSM; a stopped gradient at the actor interface decouples the two optimization paths.

\section{Experiments and Evaluations}
\label{sec:results}
To evaluate WM-Craftnet, we organize the experiments around four questions:

\begin{itemize}
    \setlength{\itemsep}{-0.5mm}
    \item \textbf{Q1. Multi-object rotation and sim-to-real:} Can WM-Craftnet learn robust multi-object rotation and transfer the learned behavior to a real robot?
    \item \textbf{Q2. WSM mechanism and prior:} What does the World Synesthesia Model improve, how does it work, and can it be reused as a synesthetic prior?
    \item \textbf{Q3. More difficult rotation modes:} Can WM-Craftnet learn rotation modes with less stable support and more complex contact rearrangement as in x-axis and y-axis rotation?
    \item \textbf{Q4. Robustness under extreme conditions:} Can WM-Craftnet rotate unseen objects and recover from challenging initial poses and external force disturbances?
\end{itemize}
\subsection{Experimental Setup}

\paragraph{Hardware and Simulation Setting.}
We use the same deployable observation/action setting and control frequency in simulation and on hardware. Each reported simulation number averages \(128\) evaluation episodes per method under randomized evaluation seeds. To reduce the sim-to-real gap, we combine joint-level system identification with domain randomization over dynamics, sensing, resets, and perturbations. Appendix~\ref{app:implementation-details} provides implementation details, and Appendix~\ref{app:sim2real} describes system identification and sim-to-real randomization.

\paragraph{Baselines.}
For the \(z\)-axis evaluation, we compare Open-loop Replay, Blind RL Policy (no depth, no tactile), Touch Dexterity (w/ tactile) \citep{yin2023touchdexterity}, RL Policy (w/ depth, w/o tactile), In-Hand Rotation (w/ depth + tactile) \citep{yuan2024robot}, and WM-Craftnet with WSM variants. The \(x\)- and \(y\)-axis evaluations use a compact baseline comparison. All baselines are reimplemented for the \(22\)-DoF Sharpa Wave Hand with the same reward, randomization range, initialization distribution, and reset strategy.

\paragraph{Metrics.}
For simulation, we report return, episode length, rotation rate (RotR), and, where informative, off-axis motion and angular-velocity variation. Simulation entries are means with \(95\%\) confidence intervals over \(128\) evaluation episodes unless otherwise noted, and highlighting marks the first three ranks under each metric direction. For hardware, RR reports radians of rotation in a \(20\)-s episode, and SR is the success rate, where success requires more than a half-spin. Four annotators independently score synchronized videos, and their measurements are averaged.

\subsection{Robust Multi-Object \(z\)-Axis Rotation}
We first evaluate whether one policy can rotate multiple everyday objects under randomized starts. The nine-object set beside Table~\ref{tab:z-axis-ablation} spans diverse shapes, mass-center distributions, textures, and contact affordances, requiring online state and contact inference rather than object-specific gaits.
As shown in Table~\ref{tab:z-axis-ablation}, baseline returns range from \(236.1\) to \(386.9\). WM-Craftnet trained from scratch improves return to \(414.3\) and reduces OffAxis and AngVar to \(1.285\) and \(1.735\), while its \(0.742\) RotR remains below some baselines. The pretrained full WSM reaches \(753.3\) return and \(1.293\) RotR, with OffAxis \(1.225\) and AngVar \(1.324\). Figure~\ref{fig:z-axis-learning-curves}(a) compares the \(z\)-axis methods; qualitative results are available on the project website\footnote{\href{https://wmcraftnet.github.io/}{\textcolor{cyanblue}{\texttt{https://wmcraftnet.github.io/}}}}.

\begin{table}[t]
  \centering
  \caption{\(z\)-axis multi-object rotation and WSM analysis on nine objects, including test-time tactile masking and modality controls.}
  \label{tab:z-axis-ablation}
  \begin{lrbox}{\zaxistablebox}
  \begin{minipage}{0.72\linewidth}
    \centering
    \begin{adjustbox}{width=\linewidth,center}
  \begin{tabular}{lccccc}
  \toprule
  Method & Return$\uparrow$ & EpLen$\uparrow$ & RotR$\uparrow$ & OffAxis$\downarrow$ & AngVar$\downarrow$ \\
  \midrule
Open-loop Replay & 306.3 $\pm$ 48.3 & 320.2 $\pm$ 35.0 & 0.332 $\pm$ 0.080 & 1.789 $\pm$ 0.295 & 2.241 $\pm$ 0.679 \\
Blind RL Policy (no depth, no tactile) & 304.1 $\pm$ 3.8 & 295.8 $\pm$ 2.3 & 0.944 $\pm$ 0.004 & 1.539 $\pm$ 0.007 & 1.810 $\pm$ 0.011 \\
Touch Dexterity (w/ tactile)\citep{yin2023touchdexterity} & 386.9 $\pm$ 3.1 & 362.5 $\pm$ 2.3 & 1.018 $\pm$ 0.004 & 1.601 $\pm$ 0.008 & 1.764 $\pm$ 0.053 \\
RL Policy (w/ depth, w/o tactile) & 362.2 $\pm$ 4.7 & 351.0 $\pm$ 3.6 & 0.978 $\pm$ 0.004 & 1.806 $\pm$ 0.009 & 2.222 $\pm$ 0.065 \\
In-Hand Rotation (w/ depth + tactile)\citep{yuan2024robot} & 236.1 $\pm$ 2.7 & 272.5 $\pm$ 2.8 & 0.882 $\pm$ 0.003 & 1.986 $\pm$ 0.009 & 2.487 $\pm$ 0.074 \\
WM-Craftnet (from scratch) & 414.3 $\pm$ 15.7 & 264.2 $\pm$ 8.8 & 0.742 $\pm$ 0.012 & 1.285 $\pm$ 0.010 & 1.735 $\pm$ 0.052 \\
\midrule
WM-Craftnet (all-zero tactile) & 724.9 $\pm$ 4.4 & 428.8 $\pm$ 2.8 & 1.252 $\pm$ 0.004 & {\nd\sl 1.224 $\pm$ 0.006} & 1.388 $\pm$ 0.031 \\
WM-Craftnet (75\% tactile dropout) & 735.2 $\pm$ 4.1 & {\rd 429.8 $\pm$ 2.7} & 1.272 $\pm$ 0.004 & {\nd\sl 1.224 $\pm$ 0.006} & 1.390 $\pm$ 0.031 \\
WM-Craftnet (50\% tactile dropout) & {\rd 737.8 $\pm$ 4.4} & 425.1 $\pm$ 3.6 & {\rd 1.276 $\pm$ 0.004} & 1.279 $\pm$ 0.006 & 1.511 $\pm$ 0.061 \\
WM-Craftnet (25\% tactile dropout) & {\nd\sl 743.4 $\pm$ 4.2} & 427.9 $\pm$ 3.2 & {\nd\sl 1.287 $\pm$ 0.004} & 1.243 $\pm$ 0.006 & 1.434 $\pm$ 0.032 \\
\midrule
WM-Craftnet (WSM pretrained, prop-only) & 684.5 $\pm$ 4.3 & 424.0 $\pm$ 2.2 & 1.191 $\pm$ 0.003 & 1.310 $\pm$ 0.007 & {\fs\bf 1.234 $\pm$ 0.013} \\
WM-Craftnet (WSM pretrained, prop+tac) & 698.9 $\pm$ 4.3 & {\fs\bf 435.4 $\pm$ 1.9} & 1.192 $\pm$ 0.002 & {\fs\bf 1.165 $\pm$ 0.006} & {\rd 1.373 $\pm$ 0.024} \\
\textbf{WM-Craftnet (full WSM)} & {\fs\bf 753.3 $\pm$ 3.6} & {\nd\sl 435.2 $\pm$ 1.9} & {\fs\bf 1.293 $\pm$ 0.004} & {\rd 1.225 $\pm$ 0.006} & {\nd\sl 1.324 $\pm$ 0.027} \\
\bottomrule
  \end{tabular}
  \end{adjustbox}
  \end{minipage}
  \end{lrbox}
  \begin{minipage}[t]{0.26\linewidth}
    \vspace{0pt}
    \centering
    \includegraphics[width=\linewidth,height=\dimexpr\ht\zaxistablebox+\dp\zaxistablebox\relax,keepaspectratio]{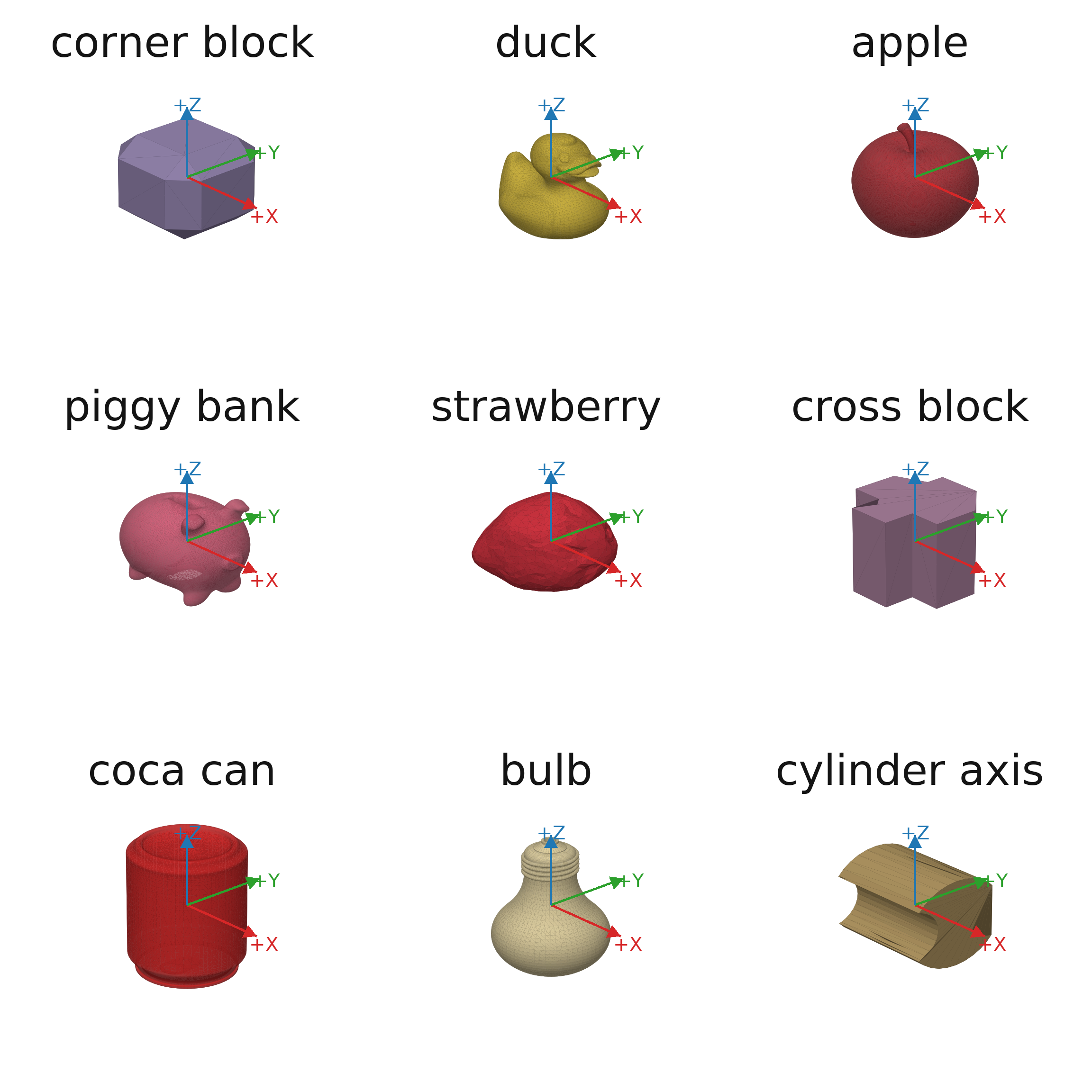}
  \end{minipage}\hfill
  \begin{minipage}[t]{0.72\linewidth}
    \vspace{0pt}
    \centering
    \usebox{\zaxistablebox}
  \end{minipage}
  \end{table}

Figure~\ref{fig:z-axis-learning-curves}(a) reports training reward for the \(z\)-axis methods and shows the same ordering as the main comparison in Table~\ref{tab:z-axis-ablation}. Table~\ref{tab:z-axis-ablation} also shows that reducing tactile availability from full input to \(25\%\) dropout and all-zero input decreases return from \(753.3\) to \(743.4\) and \(724.9\). The modality controls progress from \(684.5\) with proprioception alone to \(698.9\) with proprioception and touch, and \(753.3\) with full depth--touch input; prop+touch yields lower OffAxis (\(1.165\)) but lower RotR than full WSM. Figure~\ref{fig:z-axis-learning-curves}(b) shows these controls.

\paragraph{Controlled WSM variants.}
Table~\ref{tab:controlled-wsm-ablation} isolates clean-depth supervision, deterministic recurrence, and the recurrent policy feature. Reconstructing noisy rather than clean depth lowers return from \(753.3\) to \(708.0\), while replacing the recurrent policy context with a per-step encoder--decoder feature lowers it to \(667.6\). Removing \(h_{t-1}\) yields \(705.4\) return and \(1.282\) RotR, but improves OffAxis and AngVar to \(1.125\) and \(1.185\), indicating a stability--task-performance trade-off. The training curves beside the table show that full WSM reaches the highest final reward.

\begin{table}[t]
  \centering
  \caption{Controlled WSM variants on the \(z\)-axis benchmark. The left panel shows training reward; the right panel reports evaluation metrics (mean \(\pm\) \(95\%\) confidence interval).}
  \label{tab:controlled-wsm-ablation}
  \begin{lrbox}{\wsmcontroltablebox}
  \begin{minipage}{0.64\linewidth}
    \centering
    \scriptsize
    \setlength{\tabcolsep}{2.2pt}
    \begin{adjustbox}{width=\linewidth,center}
    \begin{tabular}{lccccc}
      \toprule
      Method & Return$\uparrow$ & EpLen$\uparrow$ & RotR$\uparrow$ & OffAxis$\downarrow$ & AngVar$\downarrow$ \\
      \midrule
      Noisy-depth supervision & {\nd\sl 708.0 $\pm$ 3.5} & 431.1 $\pm$ 2.0 & 1.265 $\pm$ 0.005 & 1.299 $\pm$ 0.006 & 1.456 $\pm$ 0.045 \\
      No \(h_{t-1}\) recurrent input & {\rd 705.4 $\pm$ 4.3} & {\nd\sl 434.3 $\pm$ 2.3} & {\nd\sl 1.282 $\pm$ 0.005} & {\fs\bf 1.125 $\pm$ 0.006} & {\fs\bf 1.185 $\pm$ 0.008} \\
      Encoder--decoder \(z_t\) policy & 667.6 $\pm$ 4.7 & 417.4 $\pm$ 2.6 & 1.254 $\pm$ 0.003 & 1.323 $\pm$ 0.006 & {\rd 1.343 $\pm$ 0.018} \\
      Full WSM & {\fs\bf 753.3 $\pm$ 3.6} & {\fs\bf 435.2 $\pm$ 1.9} & {\fs\bf 1.293 $\pm$ 0.004} & {\nd\sl 1.225 $\pm$ 0.006} & {\nd\sl 1.324 $\pm$ 0.027} \\
      \bottomrule
    \end{tabular}
    \end{adjustbox}
  \end{minipage}
  \end{lrbox}
  \begin{minipage}[c]{0.34\linewidth}
    \centering
    \includegraphics[width=\linewidth,height=\dimexpr\ht\wsmcontroltablebox+\dp\wsmcontroltablebox\relax,keepaspectratio]{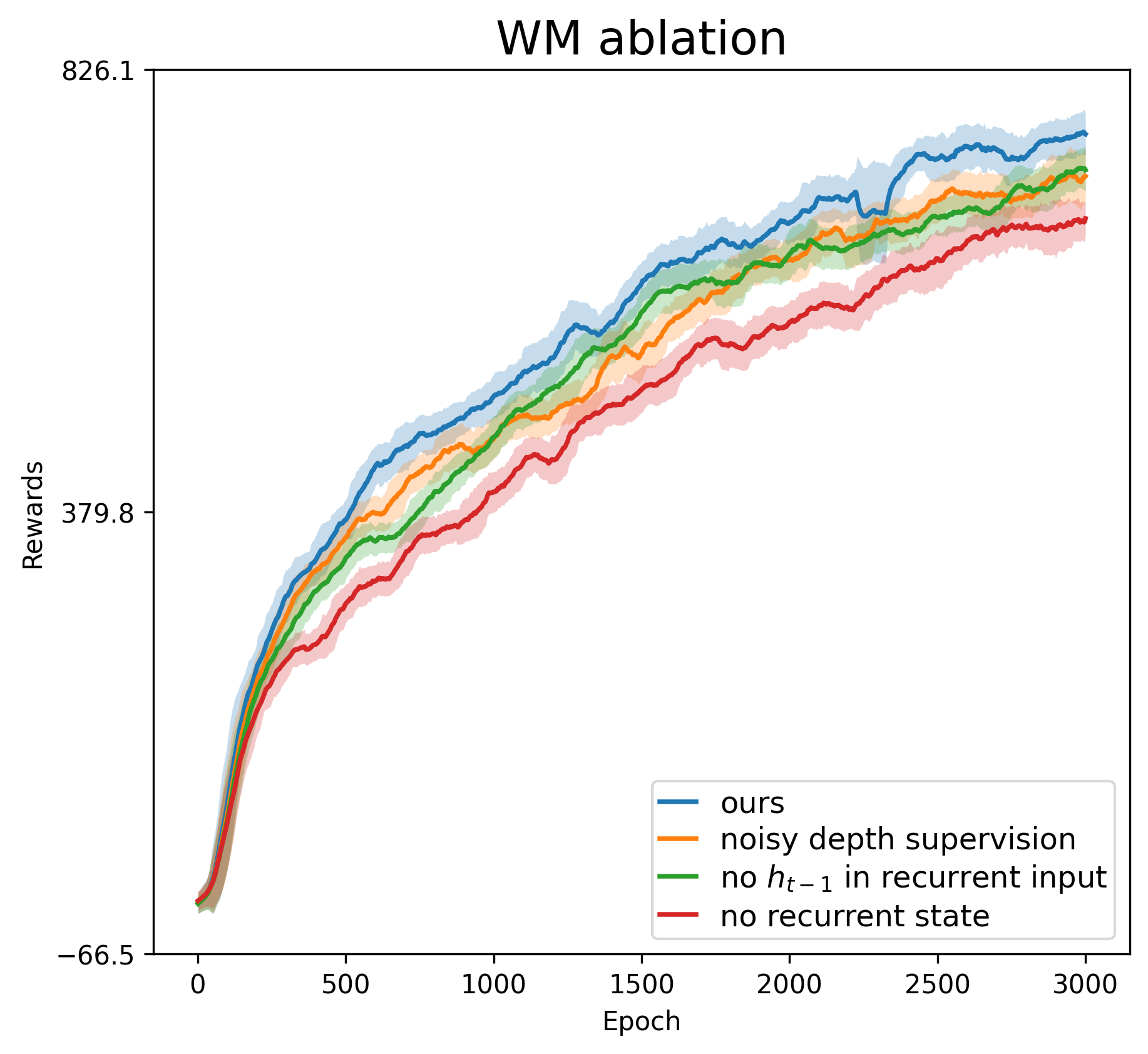}
  \end{minipage}\hfill
  \begin{minipage}[c]{0.64\linewidth}
    \centering
    \usebox{\wsmcontroltablebox}
  \end{minipage}
\end{table}

\paragraph{Sim-to-real experiments.}
We also evaluate real-robot \(z\)-axis rotation on duck, cross block, corner block, and an unseen double-notched block.
Table~\ref{tab:real-z-axis} compares a progression from fixed open-loop replay (OlR), to tactile feedback in Touch Dexterity (TD) \citep{yin2023touchdexterity}, and then depth--tactile feedback in In-Hand Rotation (IHR) \citep{yuan2024robot}. IHR is sensitive to the shift from simulated depth to noisy real observations. Feeding it WSM-denoised depth improves rotation and success on the three seen objects, confirming the value of geometric denoising, but yields limited rotation and no successful trials on the unseen block. WM-Craftnet conditions the actor on a recurrent WSM feature learned jointly from depth, touch, proprioception, and action history. This temporally informed context supports high RR and SR on both the seen objects and the unseen block.


\begin{figure}[t]
  \centering
  \begin{minipage}[t]{0.62\linewidth}
    \vspace{0pt}
    \centering
    \begin{adjustbox}{max height=0.32\textwidth,max width=\linewidth,center}
      \includegraphics{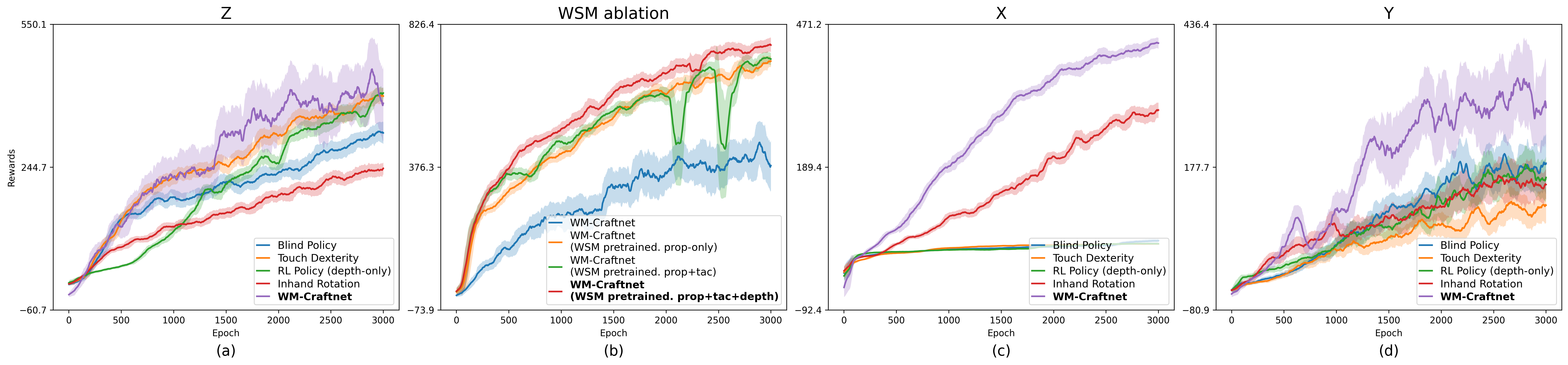}
    \end{adjustbox}
    \captionof{figure}{Training reward curves on (a) \(z\)-axis baselines, (b) WSM ablation and (c,d) \(x/y\)-axis baselines.}
    \label{fig:z-axis-learning-curves}
  \end{minipage}\hfill
  \begin{minipage}[t]{0.38\linewidth}
    \vspace{0pt}
    \centering
    {\fontsize{4.8}{5.5}\selectfont
    \setlength{\tabcolsep}{1.45pt}
    \begin{adjustbox}{max height=0.32\textwidth,max width=\linewidth,center}
    \begin{tabular}{lcccccccc}
    \toprule
    Method & \multicolumn{2}{c}{\begin{tabular}[c]{@{}c@{}}Duck\end{tabular}} & \multicolumn{2}{c}{\begin{tabular}[c]{@{}c@{}}Cross\\block\end{tabular}} & \multicolumn{2}{c}{\begin{tabular}[c]{@{}c@{}}Corner\\block\end{tabular}} & \multicolumn{2}{c}{\begin{tabular}[c]{@{}c@{}}Double-notched\\block(unseen)\end{tabular}} \\
    \cmidrule(lr){2-3}\cmidrule(lr){4-5}\cmidrule(lr){6-7}\cmidrule(lr){8-9}
     & RR & SR & RR & SR & RR & SR & RR & SR \\
    \midrule
    OlR & 2.32  & 6/10 & 3.04 & 6/10 & 5.19 & 7/10 & 0.23  & 0/10\\
    TD\citep{yin2023touchdexterity} & 1.54 & 4/10 & 0.34 & 2/10 & 0.22 & 0/10 & 0.27 & 0/10 \\
    IHR\citep{yuan2024robot} & 1.83 & 5/10& 0.45 & 1/10 & 0.36 & 1/10 & 0.39 & 0/10 \\
    IHR + WSM-denoised depth & 2.76 & 8/10 & 1.21 & 8/10 & 2.43 & 10/10 & 0.80 & 0/10 \\
    Ours & \fs\bf 16.179 &\fs\bf 10/10 &\fs\bf 8.01 &\fs\bf 10/10 &\fs\bf 8.48 &\fs\bf 10/10 &\fs\bf 4.32 &\fs\bf 8/10 \\
    \bottomrule
    \end{tabular}
    \end{adjustbox}
    }
    \captionof{table}{Real-robot \(z\)-axis rotation. RR is radians per \(20\)-s episode and SR is success over \(10\) trials.}
    \label{tab:real-z-axis}
  \end{minipage}
\end{figure}
\subsection{World Synesthesia Model Analysis}
We next analyze the representation learned by WSM. The results suggest that its recurrent context combines cues relevant to geometry, contact evolution, and object motion.

The auxiliary-head study in Table~\ref{tab:wsm-extra-ablation} raises return from \(688.1\) without auxiliary heads to \(737.0\) with the value head alone and \(753.3\) with all heads. Table~\ref{tab:controlled-wsm-ablation} provides the controlled clean-depth and recurrent-state variants, while the LSTM and GRU history baselines remain in Appendix~\ref{app:wsm-mechanisms}.

We further examine the learned state with two diagnostics. For the nine \(z\)-axis objects, we project evaluation-time recurrent states \(\mathbf{h}_t\) with t-SNE to characterize object- and interaction-phase structure; object labels are used only to color the visualization and are not policy inputs. We also decode clean depth from the latent state and compare it with ground-truth and noisy wrist-depth inputs.

\begin{figure}[t]
  \centering
  \begin{minipage}[t]{0.32\linewidth}
    \centering
    \includegraphics[height=0.6\textwidth]{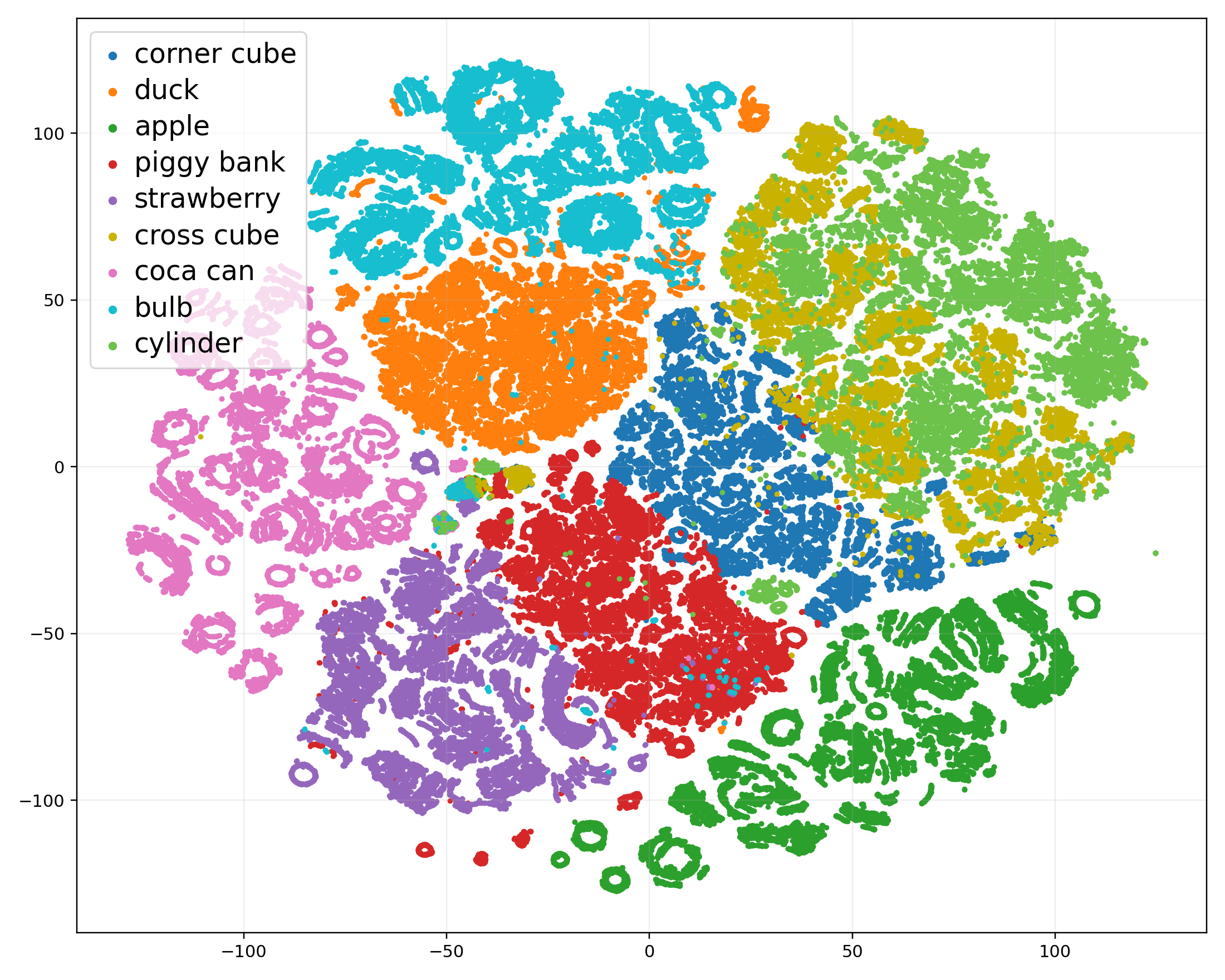}
    \caption{t-SNE visualization of WSM recurrent states \(\mathbf{h}_t\) collected from \(z\)-axis rotation rollouts over nine objects.}
    \label{fig:wm-ht-tsne}
  \end{minipage}\hfill
  \begin{minipage}[t]{0.32\linewidth}
    \centering
    \includegraphics[height=0.6\textwidth]{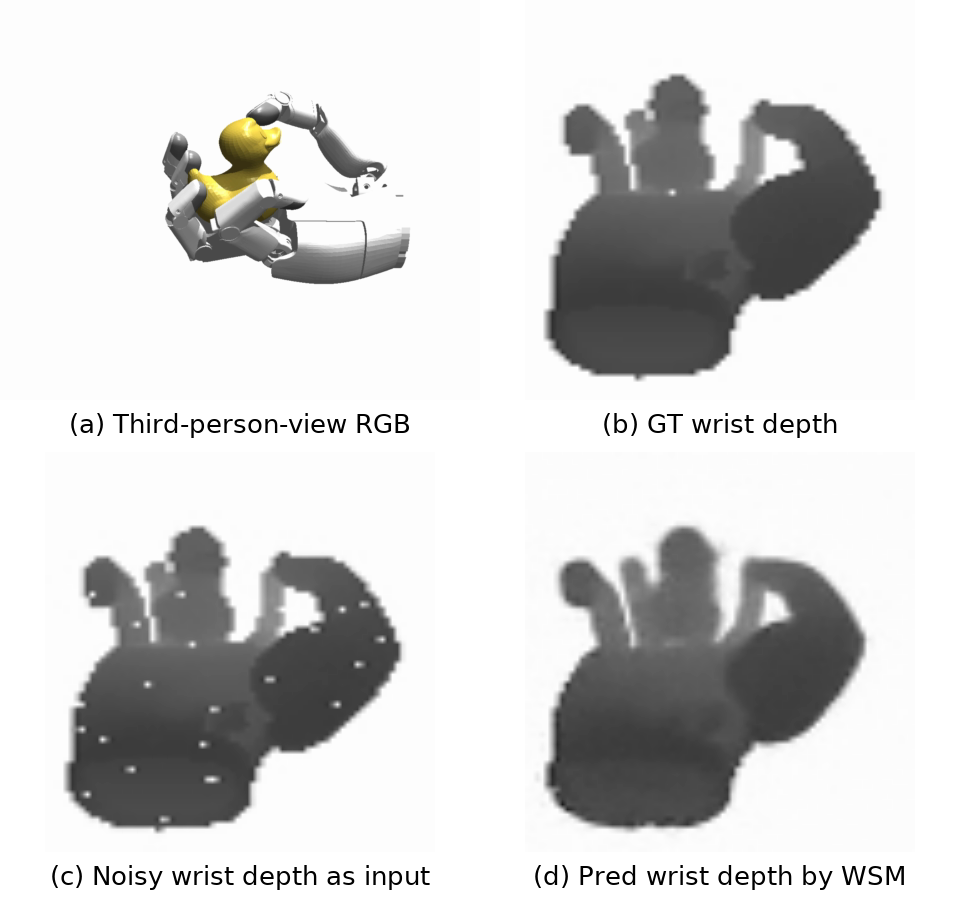}
    \caption{Depth reconstruction from a simulated \(z\)-axis duck rotation rollout. }
    \label{fig:wsm-depth-reconstruction}
  \end{minipage}\hfill
  \begin{minipage}[t]{0.32\linewidth}
    \centering
    \includegraphics[height=0.6\textwidth]{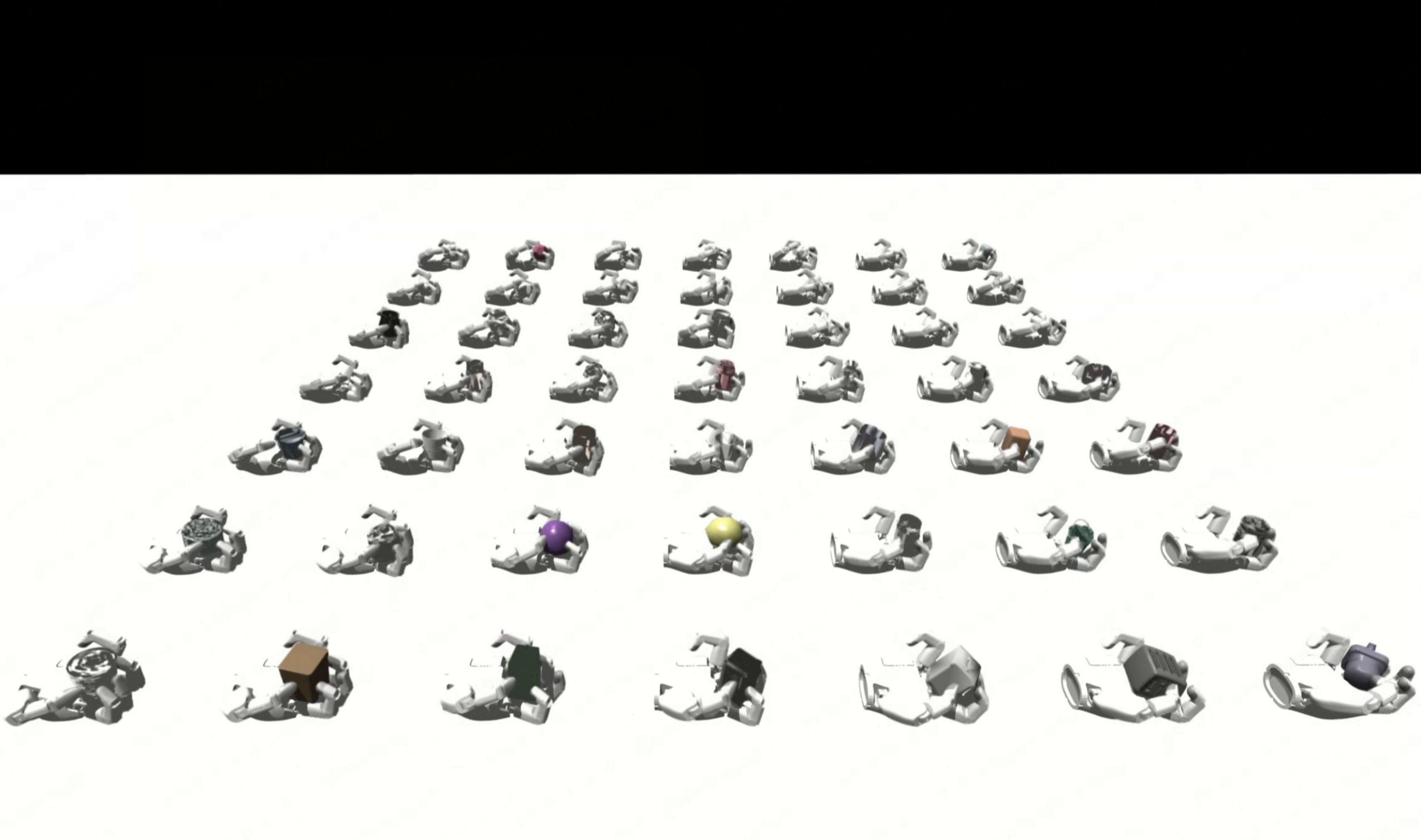}
    \caption{Downstream policy learning over \(49\) new objects using the WSM pretrained on nine \(z\)-axis objects.}
    \label{fig:multi-object-training}
  \end{minipage}
  \vspace{0mm}
\end{figure}

\textbf{Latent-state and reconstruction analysis.} Figure~\ref{fig:wm-ht-tsne} shows partially object-dependent regions together with overlap across objects and rollout phases. Figure~\ref{fig:wsm-depth-reconstruction} shows that WSM reconstructs cleaner depth from noisy input, suppressing artifacts while preserving hand-object geometry; real-hardware videos show the same effect. Together, these diagnostics are consistent with recurrent states that reflect both object-related geometry and shared interaction phases under noisy depth observations.

\paragraph{Reusable synesthetic prior.}
The same predictive model can be reused beyond the nine-object \(z\)-axis setting. Figure~\ref{fig:multi-object-training} and website videos show one downstream policy trained over \(49\) new objects; after \(3000\) epochs, each object rotates \(\boldsymbol{9.37\pm0.13}\) \textbf{rad} per episode on average, compared with \(3.28\) \textbf{rad} for the no-prior baseline, while the training fall rate drops from \(\boldsymbol{6\%}\) to \(\boldsymbol{0.3\%}\) from start to end. In a separate five-block prior-scaling diagnostic (diagonal, narrow, double-notched, step, and V-cut blocks), RotR improves from scratch / 9-object prior / 49-object prior as \(1.10/1.34/1.46\). This diagnostic is distinct from the axis-specific held-out sets below; Appendix~\ref{app:object-set} shows the full object set and Figure~\ref{fig:49-object-tsne} visualizes its recurrent-state organization.

\subsection{More Difficult Rotation Modes}
\vspace{0mm}
We next evaluate rotation axes with less passive palm support. \(y\)-axis rotation creates larger moment arms and asymmetric contacts for elongated objects; \(x\)-axis rotation is harder because the hand has limited fingertip workspace and must reallocate contacts while gravity pulls the object away from the palm. All axis-specific policies use the same WM-Craftnet design, reward, randomized starts, and deployable observations, with a separately trained WSM for each axis.

The \(y\)-axis evaluation uses the tool-like object set shown with Table~\ref{tab:y-axis}, testing whether the policy can rotate objects that tend to lever against the fingers.
Table~\ref{tab:y-axis} shows that WM-Craftnet improves return from the strongest baseline value of \(177.8\) to \(272.7\), nearly doubles the best baseline RotR from \(0.545\) to \(1.025\), and achieves the lowest fall rate. The \(88.1\%\) RotR gain is substantially larger than the \(11.7\%\) EpLen gain, indicating that the return improvement reflects faster target-axis progress rather than only longer episodes. Its ObjVel ranks second rather than first, so the advantage is specific to controlled rotation and fall reduction, not uniformly lower object motion. Figure~\ref{fig:z-axis-learning-curves}(d) shows the same training trend, suggesting that the recurrent state helps resolve the changing moment arms and asymmetric contacts.

\begin{table}[t]
\centering
\caption{\(y\)-axis rotation on nine tool-like objects. Metrics are averaged over the nine-object set shown in the adjacent grid and all evaluation trials.}
\label{tab:y-axis}
\begin{lrbox}{\yaxistablebox}
\begin{minipage}{0.70\linewidth}
\centering
\begin{adjustbox}{width=\linewidth,center}
\begin{tabular}{lccccc}
\toprule
Method & Return$\uparrow$ & EpLen$\uparrow$ & RotR$\uparrow$ & Fall$\downarrow$ & ObjVel$\downarrow$ \\
\midrule
Open-loop Replay & -52.4 $\pm$ 5.2 & 85.8 $\pm$ 10.0 & 0.031 $\pm$ 0.063 & 0.784 $\pm$ 0.082 & {\fs\bf 0.0366 $\pm$ 0.0042} \\
Blind RL Policy (no depth, no tactile) & {\nd\sl 177.8 $\pm$ 9.2} & {\nd\sl 192.4 $\pm$ 7.6} & {\nd\sl 0.545 $\pm$ 0.016} & 0.164 $\pm$ 0.002 & {\rd 0.0737 $\pm$ 0.0014} \\
Touch Dexterity (w/ tactile)\citep{yin2023touchdexterity} & 105.3 $\pm$ 5.8 & 172.0 $\pm$ 6.2 & 0.297 $\pm$ 0.030 & 0.208 $\pm$ 0.003 & 0.0760 $\pm$ 0.0021 \\
RL Policy (w/ depth, w/o tactile) & {\rd 154.8 $\pm$ 6.9} & {\rd 178.4 $\pm$ 5.4} & 0.288 $\pm$ 0.033 & {\rd 0.103 $\pm$ 0.002} & 0.0909 $\pm$ 0.0015 \\
In-Hand Rotation (w/ depth + tactile)\citep{yuan2024robot} & 142.8 $\pm$ 5.5 & 162.5 $\pm$ 4.8 & {\rd 0.377 $\pm$ 0.025} & {\nd\sl 0.063 $\pm$ 0.003} & 0.0787 $\pm$ 0.0021 \\
WM-Craftnet & {\fs\bf 272.7 $\pm$ 13.5} & {\fs\bf 215.0 $\pm$ 8.7} & {\fs\bf 1.025 $\pm$ 0.026} & {\fs\bf 0.049 $\pm$ 0.001} & {\nd\sl 0.0682 $\pm$ 0.0013} \\
\bottomrule
\end{tabular}
\end{adjustbox}
\end{minipage}
\end{lrbox}
\begin{minipage}[t]{0.26\linewidth}
  \vspace{0pt}
  \centering
  \includegraphics[height=\dimexpr\ht\yaxistablebox+\dp\yaxistablebox\relax]{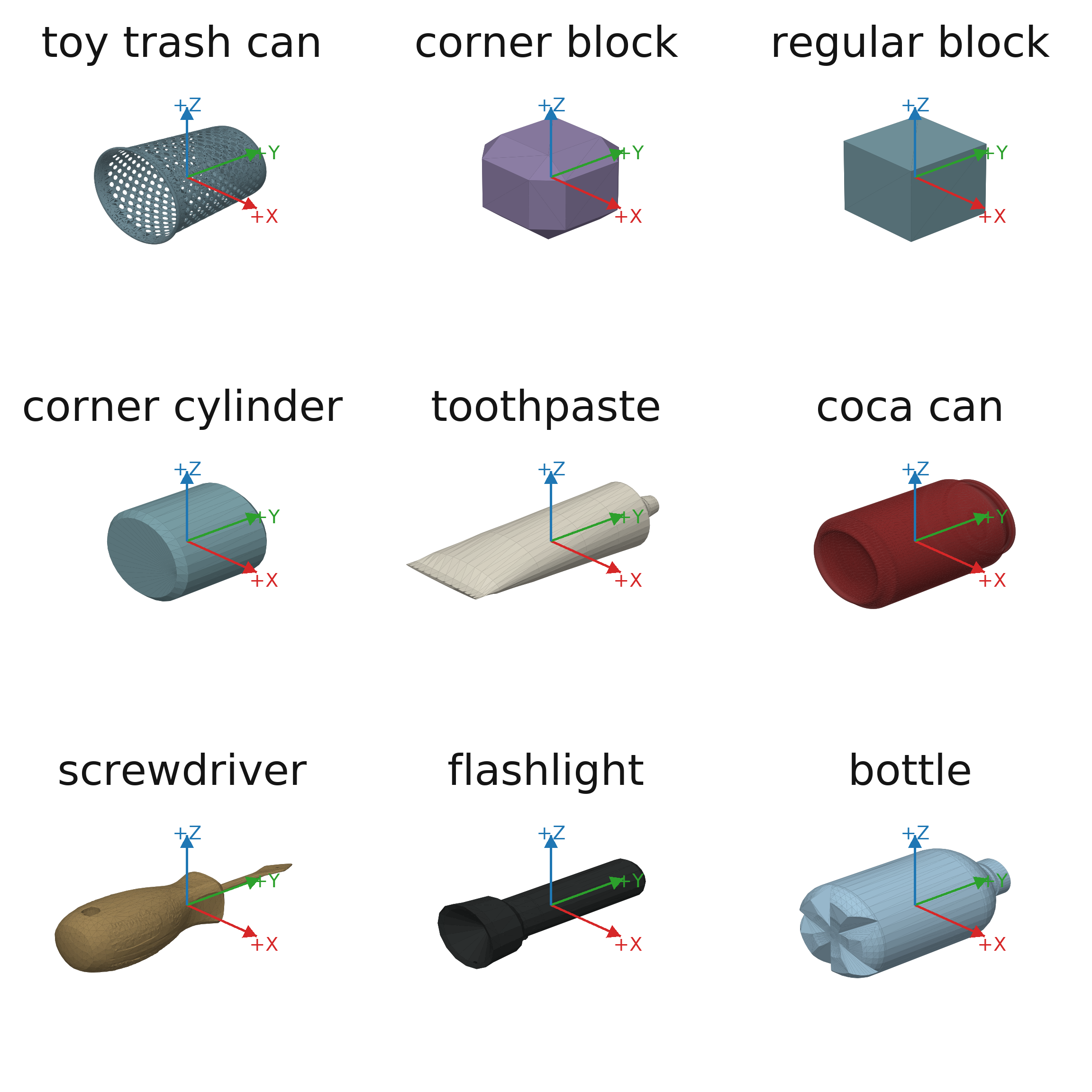}
\end{minipage}\hfill
\begin{minipage}[t]{0.70\linewidth}
  \vspace{0pt}
  \centering
  \usebox{\yaxistablebox}
\end{minipage}
\vspace{0mm}
\end{table}

The \(x\)-axis evaluation in Table~\ref{tab:x-axis} is a stricter test of contact sequencing: the policy must roll the object within a small effective finger workspace and reallocate contacts before the object leaves the palm-supported configuration.
\begin{table}[t]
\centering
\caption{\(x\)-axis rotation on four objects. Metrics are averaged over the four-object set shown in the adjacent grid and all evaluation trials.}
\vspace{0mm}
\label{tab:x-axis}
\begin{lrbox}{\xaxistablebox}
\begin{minipage}{0.70\linewidth}
\centering
\begin{adjustbox}{width=\linewidth,center}
\begin{tabular}{lccccc}
\toprule
Method & Return$\uparrow$ & EpLen$\uparrow$ & RotR$\uparrow$ & Fall$\downarrow$ & ObjVel$\downarrow$ \\
\midrule
Open-loop Replay & 18.0 $\pm$ 6.8 & {\rd 129.4 $\pm$ 22.3} & 0.411 $\pm$ 0.054 & 0.164 $\pm$ 0.064 & 0.0448 $\pm$ 0.0019 \\
Blind RL Policy (no depth, no tactile) & {\rd 44.0 $\pm$ 0.8} & 63.4 $\pm$ 1.0 & 0.867 $\pm$ 0.002 & {\rd 0.033 $\pm$ 0.001} & {\fs\bf 0.0311 $\pm$ 0.0001} \\
Touch Dexterity (w/ tactile)\citep{yin2023touchdexterity} & 37.8 $\pm$ 0.2 & 42.7 $\pm$ 0.2 & {\rd 1.023 $\pm$ 0.002} & 0.039 $\pm$ 0.002 & {\rd 0.0334 $\pm$ 0.0001} \\
RL Policy (w/ depth, w/o tactile) & 38.2 $\pm$ 0.2 & 43.7 $\pm$ 0.4 & {\nd\sl 1.079 $\pm$ 0.003} & {\nd\sl 0.033 $\pm$ 0.001} & {\nd\sl 0.0323 $\pm$ 0.0001} \\
In-Hand Rotation (w/ depth + tactile)\citep{yuan2024robot} & {\nd\sl 295.1 $\pm$ 3.3} & {\nd\sl 365.1 $\pm$ 3.1} & 0.839 $\pm$ 0.001 & 0.103 $\pm$ 0.003 & 0.0433 $\pm$ 0.0001 \\
WM-Craftnet & {\fs\bf 432.8 $\pm$ 2.1} & {\fs\bf 435.4 $\pm$ 1.8} & {\fs\bf 1.114 $\pm$ 0.001} & {\fs\bf 0.020 $\pm$ 0.001} & 0.0460 $\pm$ 0.0001 \\
\bottomrule
\end{tabular}
\end{adjustbox}
\end{minipage}
\end{lrbox}
\begin{minipage}[t]{0.26\linewidth}
  \vspace{0pt}
  \centering
  \includegraphics[height=\dimexpr\ht\xaxistablebox+\dp\xaxistablebox\relax]{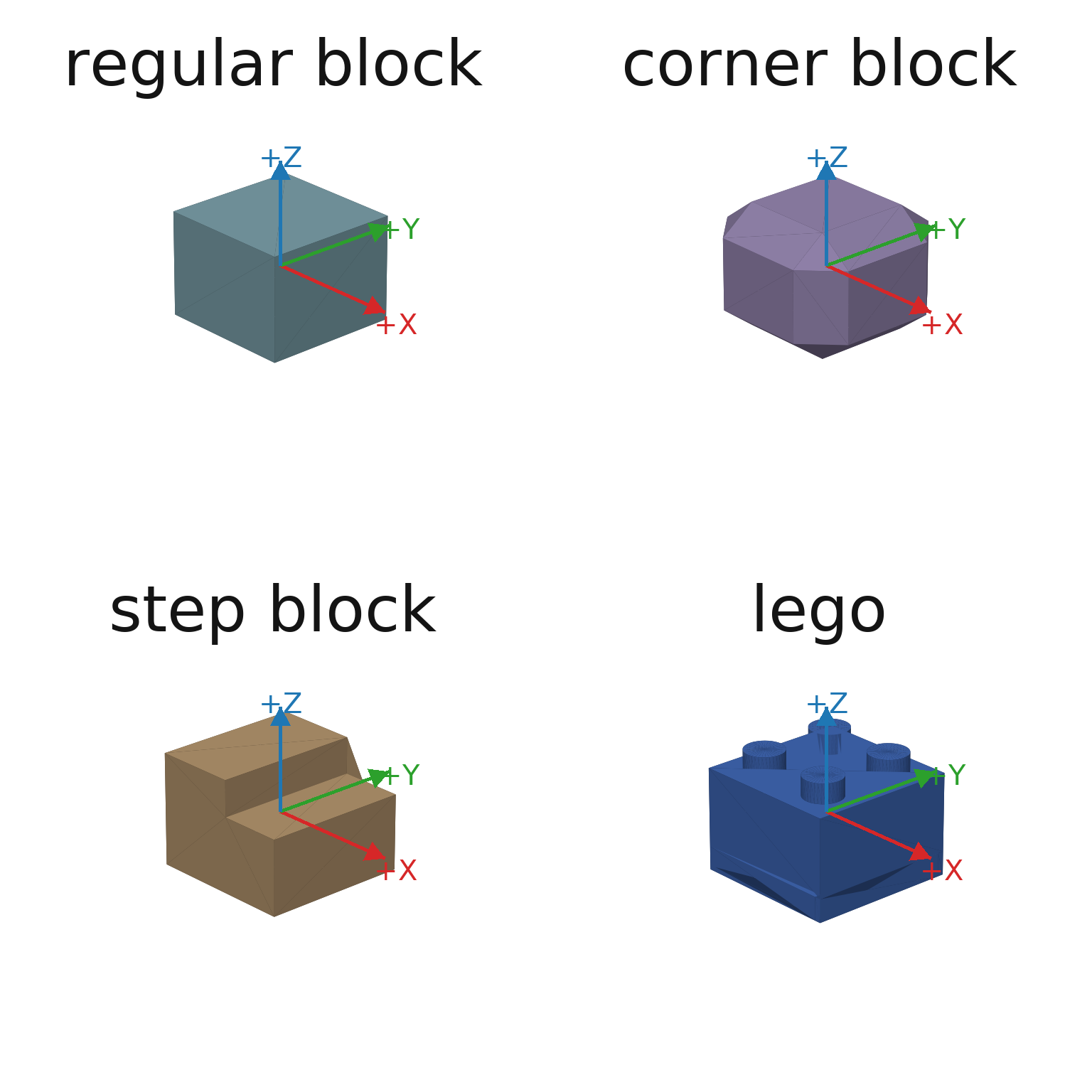}
\end{minipage}\hfill
\begin{minipage}[t]{0.70\linewidth}
  \vspace{0pt}
  \centering
  \usebox{\xaxistablebox}
\end{minipage}
\end{table}

Table~\ref{tab:x-axis} and Figure~\ref{fig:z-axis-learning-curves}(c) show that blind, tactile-only, and depth-only policies stay below \(45\) return and \(64\) episode length, while open-loop replay lasts longer but still has low return, low RotR, and a high fall rate. WM-Craftnet reaches \(432.8\) return, \(435.4\) episode length, the highest RotR, and the lowest fall rate, maintaining grasp while finding contacts that advance the commanded spin.

\subsection{Stress Tests in Challenging Interaction States}

\textbf{Unseen-object tests.}
We begin with axis-specific zero-shot transfer in simulation. We directly evaluate the \(x\)-, \(y\)-, and \(z\)-axis policies trained on their respective \(4\)-, \(9\)-, and \(9\)-object training sets, without object-specific fine-tuning. For each axis, the held-out evaluation contains four unseen objects: small corner block, narrow block, large step block, and small Lego for \(x\); double-notched block, large flashlight, narrow block, and V-cut block for \(y\); and lidded jar, tilted tumbler, handled pot, and pedal bin for \(z\). Table~\ref{tab:unseen-axis-recovery} shows that WM-Craftnet achieves the highest RotR and return in all three axis-specific evaluations. Relative to the strongest non-WSM entry on each metric, the RotR gains are \(8.6\%\), \(89.2\%\), and \(20.9\%\) for \(x\), \(y\), and \(z\), respectively; the corresponding return gains are \(35.6\%\), \(5.4\%\), and \(76.7\%\).

\textbf{Perturbation-recovery test.}
We next isolate recovery under severe simulated perturbations and measure whether the policy restores a controllable configuration, together with the elapsed time among recovered trials. WM-Craftnet reaches \(14.1\%\) recovery success, compared with \(6.2\%\) for the strongest baseline, and has the shortest mean recovery time at \(2.56\) s. This is a \(7.9\)-percentage-point absolute gain and a \(0.58\)-s reduction in mean recovery time over the strongest baseline on the respective metric. Under difficult starts and in-process force perturbations, the representative successful rollouts in Figure~\ref{fig:disturbance-recovery} show the policy re-centering, uprighting, or transferring the object back into a controllable finger workspace before resuming target-axis rotation. The absolute recovery rate remains limited under these large disturbances.

\begin{center}
  \centering
  \begin{adjustbox}{width=0.95\linewidth,center}
    \includegraphics{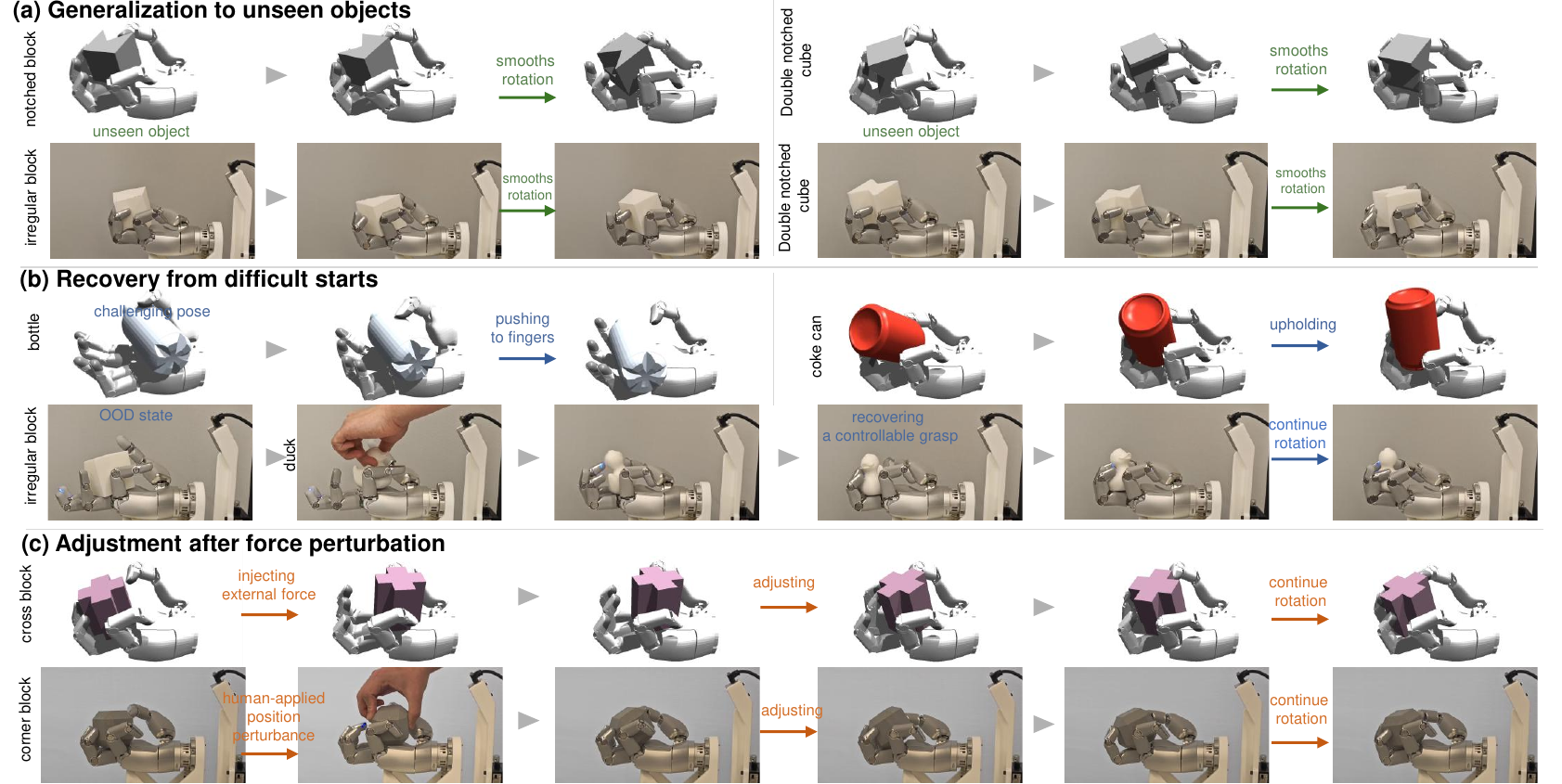}
  \end{adjustbox}
  \captionof{figure}{Qualitative stress tests in simulation and on hardware. The six rows show representative successful sequences in which WM-Craftnet returns the object to a controllable workspace and resumes rotation; quantitative recovery rates are evaluated in simulation.}
  \label{fig:disturbance-recovery}
\end{center}

\textbf{Large-scale real-world evaluation.}
Finally, in a separate large-scale hardware protocol, we test standard \(z\)-axis in-hand rotation on \(20\) real objects with \(10\) trials per object. The set combines the nine training objects with eleven unseen objects: lidded jar, tilted tumbler, pedal bin, handled pot, diagonal block, narrow block, double-notched block, step block, V-cut block, trash can, and bucket.
WM-Craftnet succeeds in \(175/200\) trials, compared with \(33/200\)--\(53/200\) for the baselines. This hardware result measures rotation success across a broad object set.

\begin{table}[t]
\centering
\caption{Stress tests on axis-specific held-out rotation, simulated \(z\)-axis perturbation recovery, and large-scale real-world \(z\)-axis rotation. The left panel shows four held-out objects per axis. Simulation entries report RotR and return; Real Rot. SR aggregates \(200\) hardware trials over \(20\) objects using the half-spin success criterion.}
\label{tab:unseen-axis-recovery}
\begin{lrbox}{\unseentablebox}
\begin{minipage}{0.70\linewidth}
\centering
\scriptsize
\setlength{\tabcolsep}{1.5pt}
\begin{adjustbox}{width=\linewidth,center}
\begin{tabular}{lcccccc}
\toprule
Method & RotR, Ret. (\(x\))$\uparrow$ & RotR, Ret. (\(y\))$\uparrow$ & RotR, Ret. (\(z\))$\uparrow$ & Sim. Recovery$\uparrow$ & Sim. Time$\downarrow$ (s) & Real Rot. SR$\uparrow$ \\
\midrule
Blind RL Policy & 0.852, 134.7 & 0.300, 135.7 & 0.665, 277.1 & 4.7\% $\pm$ 3.7\% & 3.62 $\pm$ 2.25 & 41/200 \\
Touch Dexterity \citep{yin2023touchdexterity} & {\nd\sl 0.923, 35.8} & 0.174, 78.7 & 0.605, 242.2 & {\nd\sl 6.2\% $\pm$ 4.2\%} & 3.33 $\pm$ 1.41 & 33/200 \\
RL Policy (w/o tactile) & 0.853, 33.8 & {\nd\sl 0.378, 168.9} & {\nd\sl 0.696, 301.7} & 5.5\% $\pm$ 4.0\% & {\nd\sl 3.14 $\pm$ 0.79} & 45/200 \\
In-Hand Rotation \citep{yuan2024robot}& 0.732, 246.0 & 0.309, 193.0 & 0.468, 156.1 & 0.0\% $\pm$ 0.0\% & N/A & {\nd\sl 53/200} \\
\textbf{WM-Craftnet} & {\fs\bf 1.002, 333.6} & {\fs\bf 0.715, 203.5} & {\fs\bf 0.842, 533.1} & {\fs\bf 14.1\% $\pm$ 6.0\%} & {\fs\bf 2.56 $\pm$ 0.77} & {\fs\bf 175/200} \\
\bottomrule
\end{tabular}
\end{adjustbox}
\end{minipage}
\end{lrbox}
\begin{minipage}[t]{0.28\linewidth}
  \vspace{0pt}
  \centering
  \includegraphics[height=\dimexpr\ht\unseentablebox+\dp\unseentablebox\relax]{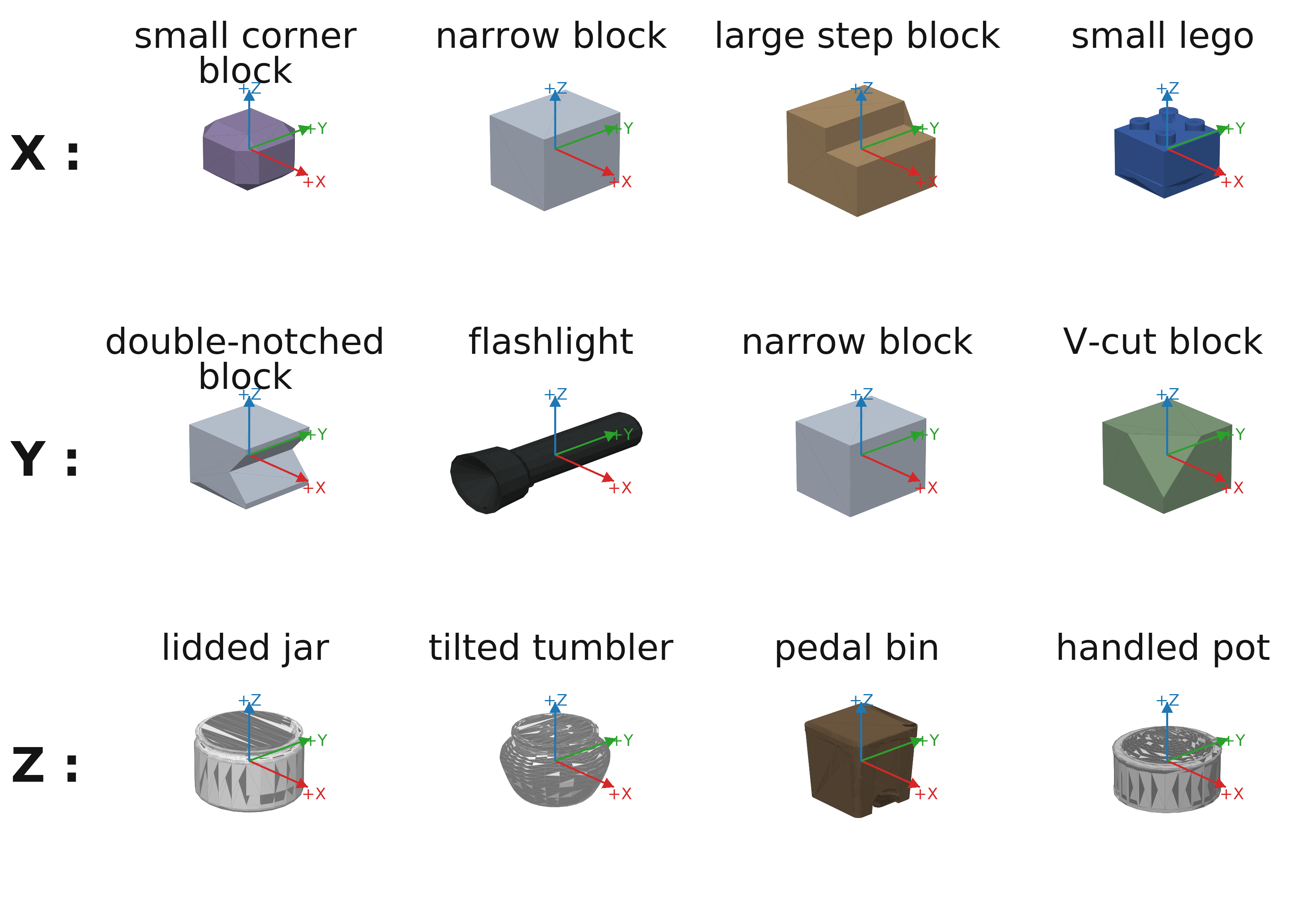}
\end{minipage}\hfill
\begin{minipage}[t]{0.70\linewidth}
  \vspace{0pt}
  \centering
  \usebox{\unseentablebox}
\end{minipage}
\end{table}

\paragraph{Tool use behavior.}
Appendix~\ref{app:tool-use} reports screwdriver translation and rotation using the same policy formulation.

\section{Conclusions and Limitations}
\label{sec:conclusion}

This paper presents WM-Craftnet, a world model-based visuotactile framework for robust dexterous in-hand manipulation. WSM learns Dreamer-style latent dynamics from noisy depth, proprioception, tactile sensing, and actions, reconstructs clean depth targets, and conditions the policy with its deterministic recurrent feature.
Depth reconstruction suppresses sensor artifacts while preserving hand-object geometry. The predictive representation also serves as a reusable prior for a \(49\)-object downstream policy. Experiments show improved multi-object rotation, unseen-object transfer, perturbation recovery, and sim-to-real performance over baseline visuotactile methods.

\textbf{Limitations:} current tasks remain short-horizon, and severe-perturbation recovery remains limited. Failures occur when objects exceed the hand workspace, large offsets leave the recovery basin, or drift and jamming disrupt stable contacts. Future work will explore broader distributions and more complex tool use.


\clearpage


\bibliography{example}  


\appendix
\newpage
\suppressfloats[t]
\section{Implementation Details}
\label{app:implementation-details}

We provide implementation details for reproducing WM-Craftnet, including the robot/simulator interface, hyper-parameters, deployable observations, reward design, and sim-to-real randomization.

\begin{table}[t]
\centering
\small
\setlength{\tabcolsep}{3pt}
\caption{Compact implementation parameters used in the current WM-Craftnet configuration.}
\vspace{-3mm}
\label{tab:implementation-params}
\begin{adjustbox}{width=0.6\linewidth,center}
\begin{tabular}{p{0.50\linewidth}p{0.40\linewidth}}
\toprule
\textbf{Parameter} & \textbf{Value} \\
\midrule
\multicolumn{2}{l}{\textbf{System}} \\
Hand & Sharpa Wave, \(22\)-DoF five-finger hand \\
Hardware rates & hand control / wrist camera / tactile: \(10\) Hz \\
Camera stream & wrist-mounted, \(640\times480\), fixed by 3D-printed bracket \\
Simulation rates & physics \(60\) Hz; observation/control \(10\) Hz \\
\midrule
\multicolumn{2}{l}{\textbf{World Model}} \\
Type & Dreamer-style RSSM with multimodal inputs \\
Deterministic / stochastic / discrete dims & \(512 / 32 / 32\) \\
Hidden size / recurrent depth & \(512 / 1\) \\
Actor projection & \(512\rightarrow64\rightarrow32\rightarrow16\) \\
Replay capacity / warmup & \(256\) rows / \(10{,}000\) transitions \\
Batch size / chunk length / updates & \(16 / 32 / 10\) per PPO epoch \\
Optimizer / LR / grad clip & Adam / \(10^{-4}\) / \(1000\) \\
KL free / dyn scale / rep scale & \(1.0 / 0.5 / 0.1\) \\
\midrule
\multicolumn{2}{l}{\textbf{Depth Input}} \\
Policy depth tensor / max depth & \(96\times72 / 0.4\) m \\
Temporal dropout / corr. & \(0.02 / 0.95\) \\
Random image rotation & \(\pm 0.3^\circ\) \\
\midrule
\multicolumn{2}{l}{\textbf{Policy and Control}} \\
Envs / horizon / minibatch / mini-epochs & \(1024 / 16 / 4096 / 4\) \\
\(\gamma\) / GAE \(\tau\) / LR / clip & \(0.99 / 0.95 / 10^{-4} / 0.2\) \\
Policy MLP / activation & \([512,256,256]\) / ELU \\
DoF / episode / sim dt / control decimation & \(22 / 500 / 0.01667 / 6\) \\
\midrule
\multicolumn{2}{l}{\textbf{Reward}} \\
Spin / max spin / velocity & \(1.0 / 2.0 / -0.1\) \\
Torque / work / action / control & \(-0.003 / -0.003 / -0.002 / -1.0\) \\
Finger distance / hand pose / off-axis & \(0.1 / 1.5 / 0.001\) \\
Object drift / no-spin / fall penalty & \(100.0 / -50.0 / -50.0\) \\
Tip force coef. / thresh. / exp. & \(0.9 / 2.0 / 0.01\) \\
\midrule
\multicolumn{2}{l}{\textbf{Randomization}} \\
Object mass train / test & \([0.2,0.4]\) kg / \([0.1,1.4]\) kg \\
Friction scale / PD scale & \([0.5,2.0] / [0.8,1.2]\) \\
Obs noise / action noise & \([0,0.05] / [0,0.04]\) \\
Reset pos. / reset rot. & \([0.035,0.02,0.0025]\) m / \(10^\circ\) \\
External force scale / prob. / decay & \(0.5 / 0.25 / 0.99\) \\
Gravity curriculum & \(1^\circ\rightarrow4^\circ\), step \(0.05^\circ\) \\
\bottomrule
\end{tabular}
\end{adjustbox}
\vspace{-1mm}
\end{table}

\begin{table}[t]
\centering
\small
\setlength{\tabcolsep}{3pt}
\caption{Deployable and privileged information used by WM-Craftnet. The actor is restricted to deployable signals; simulator-only state is used for reward, critic training, and diagnostics.}
\vspace{-1.5mm}
\label{tab:obs-action-interface}
\begin{adjustbox}{width=0.7\linewidth,center}
\begin{tabular}{p{0.28\linewidth}p{0.22\linewidth}p{0.40\linewidth}}
\toprule
\textbf{Signal} & \textbf{Used by} & \textbf{Description} \\
\midrule
Hand proprioception & Actor, WM & normalized \(22\)-DoF joint state and previous target history \\
Tactile/contact & Actor, WM & binary contact stream with threshold, dropout, and latency noise \\
Depth image & Actor, WM & noisy wrist depth input; clean depth is used only as WM reconstruction target \\
Commanded axis & Actor & goal direction \(\mathbf{a}\in\{\pm x,\pm y,\pm z\}\) \\
World-model feature & Actor & projected deterministic RSSM feature \(\psi(\mathbf{h}_t)\) \\
Relative action & Controller, WM & \(22\)-DoF joint target increment, smoothed before execution \\
Object state & Reward/critic only & pose, velocity, contact force, and fall/reset diagnostics in simulation \\
\bottomrule
\end{tabular}
\end{adjustbox}
\vspace{-1mm}
\end{table}

\subsection{World Model Training Details}
\label{app:world-model}

Table~\ref{tab:implementation-params} summarizes the world-model and depth-preprocessing settings. The world-model update receives a single-step proprioceptive vector, tactile/contact vector, noisy preprocessed depth image, previous action, and an episode-start flag; reward is stored only as a prediction target. The RSSM performs an observation update at every control step and exposes only the detached deterministic feature \(\mathbf{h}_t\), projected by a small MLP, as actor context.

World-model data are collected online from PPO rollouts into a ring replay buffer storing proprioception, tactile/contact, noisy depth input, clean crop-only target depth, previous action, reward, and episode-start flag. After warmup, contiguous chunks train clean-depth reconstruction, proprioception reconstruction, tactile-contact prediction, reward prediction, and Dreamer-style dynamics/representation KL terms. The full WSM additionally predicts object pose, value, and object shape as training objectives; the shape target is a basis-point-set-to-mesh displacement vector. These heads provide additional supervision for object state, shape, and contact-related dynamics, and are used during both pretraining and downstream adaptation.

For pretrained-prior evaluations, we transfer the encoder, multimodal RSSM, depth/proprioceptive decoder, and reward predictor. During downstream \(49\)-object PPO training, these components continue to adapt with new rollout data, while the actor--critic and task-specific heads are learned for the target distribution. This design transfers predictive physical structure across object sets while learning control for the new task distribution.

\subsection{Policy Training Details}
\label{app:policy-training}

Table~\ref{tab:implementation-params} lists PPO, control, and reward settings, and Table~\ref{tab:obs-action-interface} summarizes deployable and simulation-side signals. The actor outputs a relative target update for all \(22\) hand joints. A signal-level filter smooths these targets before execution by the \(10\) Hz hardware controller.

The reward favors stable, goal-directed, human-like rotation rather than speed alone:
\begin{equation}
\begin{aligned}
r_t
:=\;&
\lambda_{\text{spin}} r_t^{\text{spin}}
+\lambda_{\text{vel}} r_t^{\text{vel}}
+\lambda_{\text{contact}} r_t^{\text{contact}}
+\lambda_{\text{dist}} r_t^{\text{dist}}
+\lambda_{\tau} r_t^{\tau}
+\lambda_{\text{work}} r_t^{\text{work}} \\
&+
\lambda_a r_t^{a}
+\lambda_{\text{ctrl}} r_t^{\text{ctrl}}
+\lambda_{\text{pose}} r_t^{\text{pose}}
+r_t^{\text{drift}}
+r_t^{\text{axis}}
+r_t^{\text{nospin}}.
\end{aligned}
\end{equation}
The target-axis spin term uses signed incremental rotation projected onto \(\mathbf{a}\), clipped by a maximum spin rate. Stability terms penalize object velocity, drift, off-axis rotation, high torque, mechanical work, large actions, and tracking error; contact and pose terms encourage useful fingertip proximity, avoid excessive force or non-tip contacts, and softly favor natural hand postures.

Episodes reset when the object drops beyond the fall-distance threshold, deviates too far from the desired rotation axis, remains below the target-axis spin threshold for too many consecutive steps, or reaches the horizon. These resets reduce uninformative exploration after irrecoverable failures and keep all baselines under the same stability requirements.

\subsection{System Identification and Domain Randomization}
\label{app:sim2real}

Real-world evaluation uses the Sharpa Wave hand with tactile sensing and a wrist-mounted depth camera (RealSense L515). The hand controller, wrist camera, and tactile stream run at \(10\) Hz, and simulation uses the same policy/control rate with a \(60\) Hz physics step. The policy receives the same deployable observation/action interface in both domains. 

\begin{figure}[t]
  \centering
  \begin{adjustbox}{width=\linewidth,center}
    \includegraphics{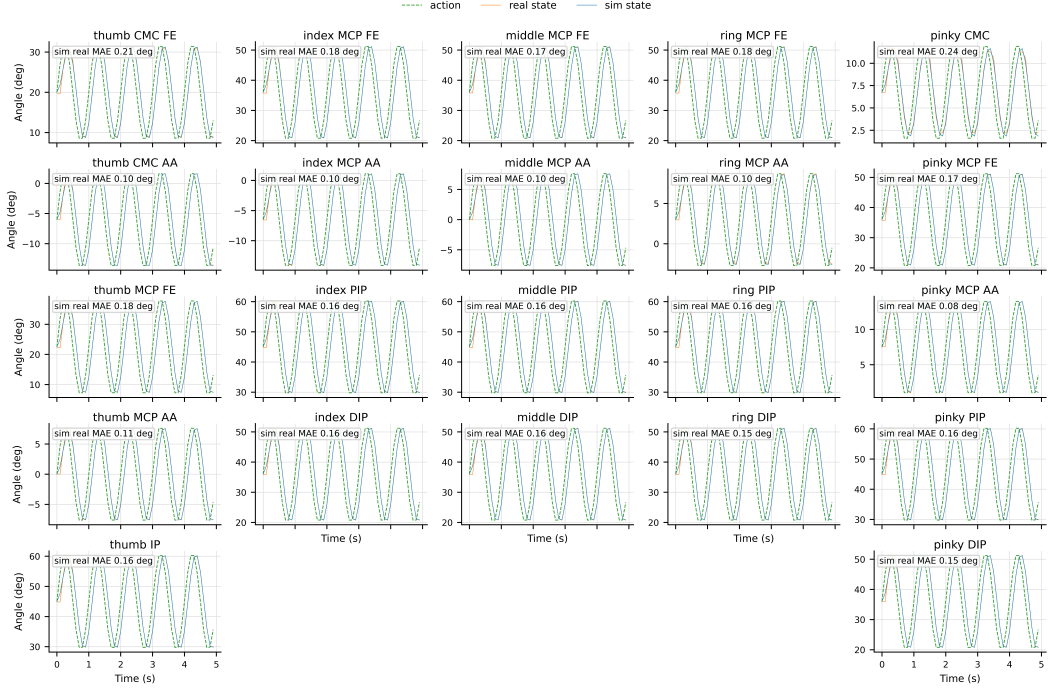}
  \end{adjustbox}
  \caption{Joint-level system identification. Dashed green curves show commanded actions, orange curves show real joint states, and blue curves show calibrated simulation states. Across the evaluated joints, the average sim-real tracking error is below \(0.2^\circ\).}
  \label{fig:pd-fit}
\end{figure}

We identify joint-level dynamics before policy transfer by exciting the real hand with \(1\) Hz sinusoidal action commands and comparing the measured joint response with simulation. The simulator \(K_p\) and \(K_d\) gains are adjusted to align the amplitude and phase of the simulated trajectories with the hardware response. This calibration provides a nominal dynamics model, while domain randomization covers residual mismatch.
Figure~\ref{fig:pd-fit} visualizes the post-calibration joint tracking fit across the hand. The simulated trajectories closely match the real joint responses under the same sinusoidal commands, with an average sim-real tracking error below \(0.2^\circ\).

Training randomizes object mass, friction, PD gains, observation/action noise, reset pose, external forces, and gravity direction as in Table~\ref{tab:implementation-params}. Object mass is sampled in \([0.2,0.4]\) kg during training and evaluated over a wider test range. Friction and hand-joint gains are randomized to cover contact and actuator uncertainty; depth and tactile observations include dropout, noise, small image rotations, threshold variation, and latency. External forces and a gravity-direction curriculum further encourage perturbation recovery instead of memorizing a single palm-up rotation gait.

\section{Additional Experiments}
\label{app:additional-experiments}

We include controlled WSM ablations, large-scale object generalization, and tool-use-style manipulation.

\subsection{Additional WSM Mechanism Ablations}
\label{app:wsm-mechanisms}

Table~\ref{tab:supplementary-wsm-ablation} compares recurrent history with the predictive WSM objective. LSTM and GRU receive the same observation histories but are trained without multimodal prediction losses.

\begin{table}[t]
  \centering
  \caption{Recurrent-history baselines and Full WSM on the \(z\)-axis benchmark (mean \(\pm\) \(95\%\) confidence interval).}
  \label{tab:supplementary-wsm-ablation}
  \vspace{-1mm}
  \small
  \setlength{\tabcolsep}{3.2pt}
  \begin{adjustbox}{width=\linewidth,center}
  \begin{tabular}{lccccc}
  \toprule
  Method & Return$\uparrow$ & EpLen$\uparrow$ & RotR$\uparrow$ & OffAxis$\downarrow$ & AngVar$\downarrow$ \\
  \midrule
  LSTM (w/o predictive obj.) & 621.0 $\pm$ 4.3 & 424.3 $\pm$ 2.7 & 1.186 $\pm$ 0.004 & 1.297 $\pm$ 0.006 & 1.392 $\pm$ 0.028 \\
  GRU (w/o predictive obj.) & 497.8 $\pm$ 4.5 & 377.1 $\pm$ 2.5 & 1.095 $\pm$ 0.003 & 1.226 $\pm$ 0.007 & 1.357 $\pm$ 0.013 \\
  \textbf{Full WSM} & {\fs\bf 753.3 $\pm$ 3.6} & {\fs\bf 435.2 $\pm$ 1.9} & {\fs\bf 1.293 $\pm$ 0.004} & {\fs\bf 1.225 $\pm$ 0.006} & {\fs\bf 1.324 $\pm$ 0.027} \\
  \bottomrule
  \end{tabular}
  \end{adjustbox}
  \vspace{-1mm}
\end{table}

Full WSM reaches \(753.3\) return and \(1.293\) RotR, compared with \(621.0/1.186\) for LSTM and \(497.8/1.095\) for GRU. It also lowers AngVar to \(1.324\), while its OffAxis \(1.225\) is comparable to GRU's \(1.226\). These results separate the benefit of predictive multimodal training from recurrent memory alone.

Table~\ref{tab:wsm-extra-ablation} measures how simulation-side auxiliary objectives shape WSM. Removing all auxiliary heads yields \(688.1\) return, while value, pose, and object-shape prediction each improve different aspects of the learned state. The value head produces the largest return gain among the single-head variants, whereas pose prediction gives the lowest angular-velocity variation. Combining all heads achieves the strongest overall return and rotation rate.

\begin{center}
\centering
\small
\setlength{\tabcolsep}{2.2pt}
\captionof{table}{Auxiliary-head ablations on the \(z\)-axis benchmark.}
\vspace{-2mm}
\label{tab:wsm-extra-ablation}
\begin{adjustbox}{width=\linewidth,center}
\begin{tabular}{lccccc}
\toprule
Method & Return$\uparrow$ & EpLen$\uparrow$ & RotR$\uparrow$ & OffAxis$\downarrow$ & AngVar$\downarrow$ \\
\midrule
No auxiliary heads & 688.1 $\pm$ 3.7 & {\rd 433.7 $\pm$ 2.0} & 1.239 $\pm$ 0.006 & {\rd 1.229 $\pm$ 0.005} & 1.778 $\pm$ 0.138 \\
Only value head & {\nd\sl 737.0 $\pm$ 4.9} & {\fs\bf 435.7 $\pm$ 2.2} & {\nd\sl 1.291 $\pm$ 0.003} & 1.258 $\pm$ 0.007 & {\rd 1.346 $\pm$ 0.045} \\
Only pose head & {\rd 690.0 $\pm$ 5.8} & 432.3 $\pm$ 2.5 & 1.221 $\pm$ 0.007 & {\nd\sl 1.228 $\pm$ 0.008} & {\fs\bf 1.231 $\pm$ 0.011} \\
Only object-shape head & 689.2 $\pm$ 4.2 & 428.1 $\pm$ 2.2 & {\rd 1.261 $\pm$ 0.002} & 1.255 $\pm$ 0.006 & 1.535 $\pm$ 0.032 \\
\midrule
\textbf{Full WSM} & {\fs\bf 753.3 $\pm$ 3.6} & {\nd\sl 435.2 $\pm$ 1.9} & {\fs\bf 1.293 $\pm$ 0.004} & {\fs\bf 1.225 $\pm$ 0.006} & {\nd\sl 1.324 $\pm$ 0.027} \\
\bottomrule
\end{tabular}
\end{adjustbox}
\vspace{-1mm}
\end{center}

\subsection{Large-Scale Object Set}
\label{app:object-set}

Figure~\ref{fig:49-object-set}(a) shows the new \(49\)-object set used for downstream large-scale multi-object policy training after WSM pretraining on the nine \(z\)-axis objects. The set covers substantial variation in size, aspect ratio, curvature, and contact geometry. To examine whether the transferred WSM retains structured interaction states at this larger scale, we project evaluation-time recurrent states from the downstream policy with t-SNE. Figure~\ref{fig:49-object-tsne}(b) shows locally coherent trajectories and partially object-dependent regions together with substantial overlap. This pattern is consistent with a representation that captures both object-dependent geometry and interaction phases shared across objects, rather than acting as a strict object-identity classifier.

\begin{center}
  \centering
  \begin{minipage}[c]{0.44\linewidth}
    \centering
    \includegraphics[height=5.3cm,keepaspectratio]{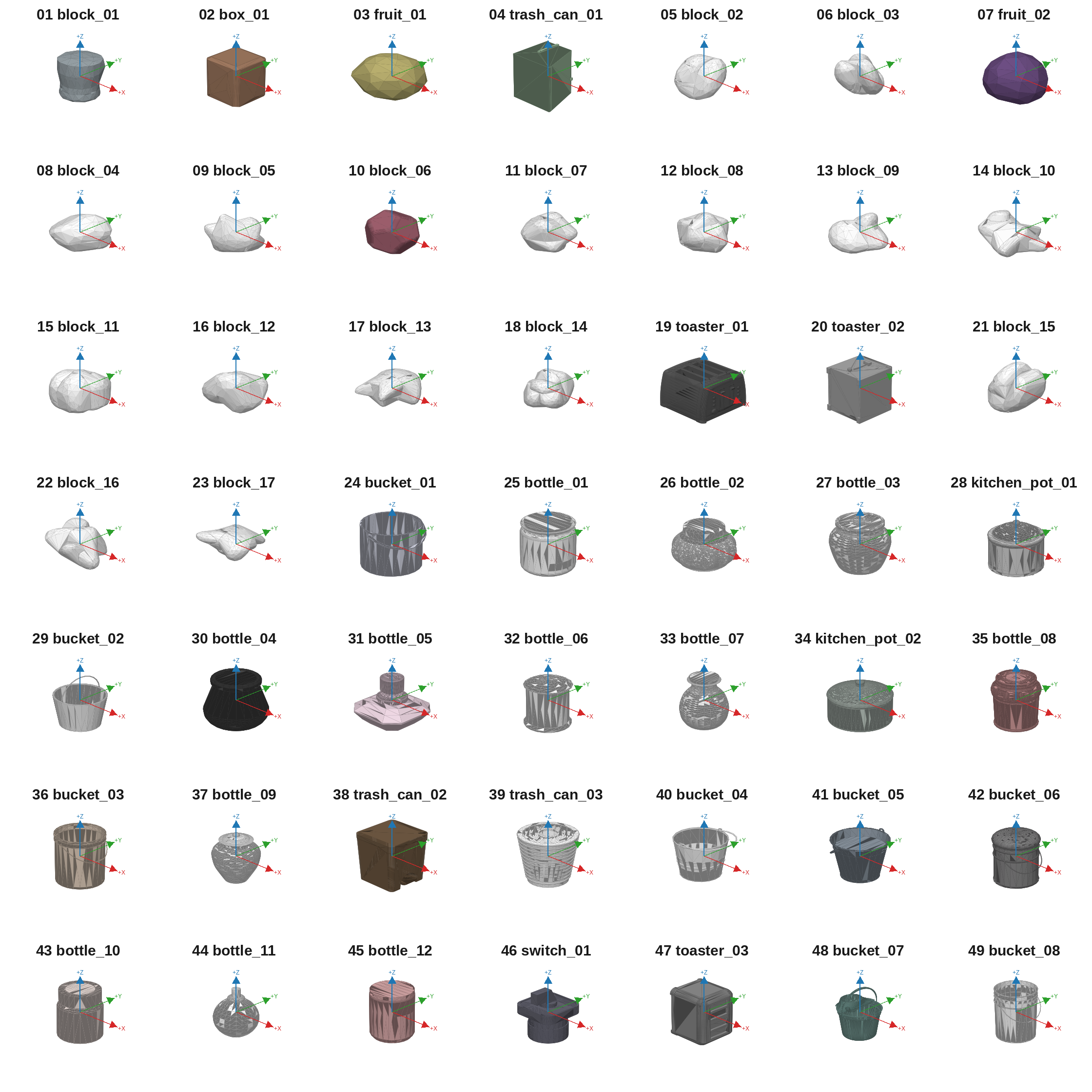}\\[-1mm]
    \textbf{(a)}
  \end{minipage}\hfill
  \begin{minipage}[c]{0.54\linewidth}
    \centering
    \includegraphics[height=5.3cm,keepaspectratio]{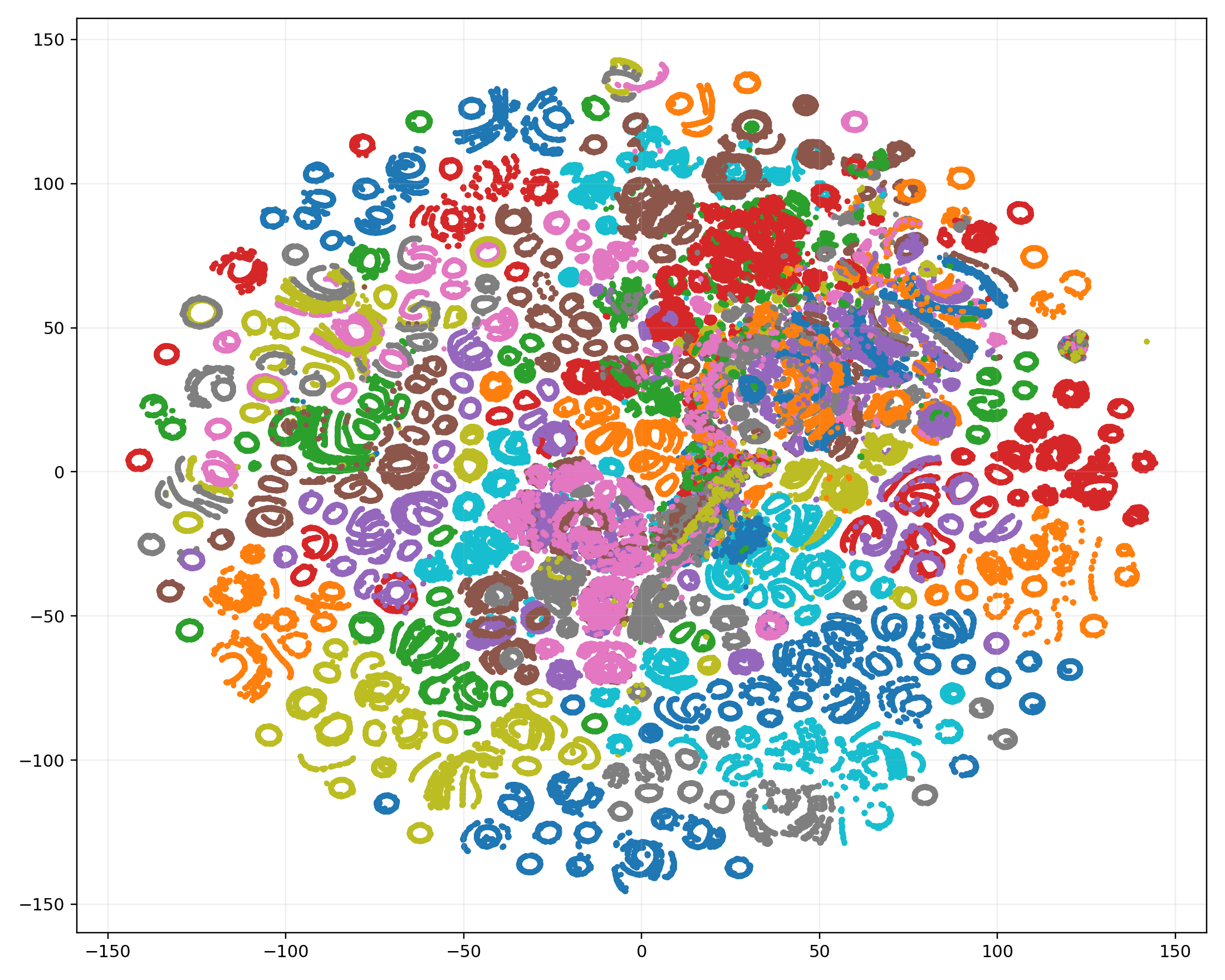}\\[-1mm]
    \textbf{(b)}
  \end{minipage}
  \captionof{figure}{Large-scale downstream analysis. (a) The \(49\)-object set used for multi-object policy training. (b) t-SNE projection of evaluation-time WSM recurrent states, with colors denoting object identities; locally coherent trajectories, partially separated regions, and shared regions coexist across the distribution.}
  \label{fig:49-object-set}
  \label{fig:49-object-tsne}
\end{center}

\subsection{Screwdriver Tool-Use Manipulation}
\label{app:tool-use}

We also examine screwdriver manipulation beyond benchmark rotation, where elongated geometry, local handle contacts, slip risk, and task-dependent adjustment test tool-use-style behavior. Figure~\ref{fig:tool-use} shows that the same WM-Craftnet interface can adjust the screwdriver to a target position in goal-conditioned translation (a) and rotate it while maintaining contact (b).

\begin{center}
  \centering
  \begin{adjustbox}{width=0.55\linewidth,center}
    \includegraphics{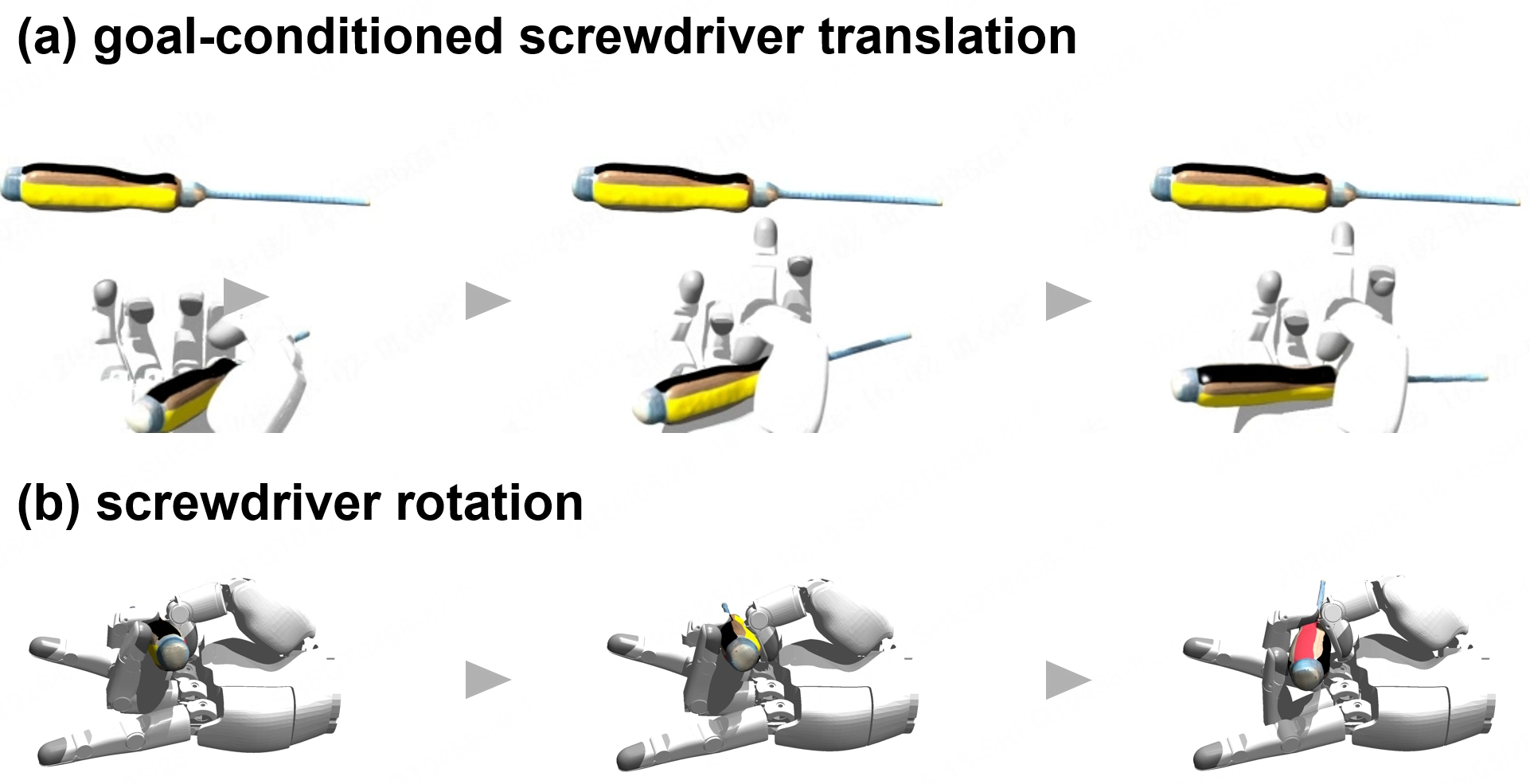}
  \end{adjustbox}
  \captionof{figure}{Screwdriver tool-use manipulation. (a) Goal-conditioned screwdriver translation, where the policy adjusts the screwdriver to the target position. (b) Screwdriver rotation, where the policy rotates an elongated screwdriver object.}
  \label{fig:tool-use}
\end{center}

\subsection{Supplementary rollouts and failure cases.}
Our project website provides supplementary real-world rollouts, qualitative comparisons, and failure cases involving depth noise, slip, and difficult contact transitions.


\end{document}